\pdfoutput=1

\documentclass[11pt]{article}

\usepackage[final]{acl}

\usepackage{times}
\usepackage{latexsym}
\usepackage{subcaption} 
\usepackage{booktabs}
\usepackage{amsmath}
\usepackage{makecell}
\usepackage[T1]{fontenc}

\usepackage[utf8]{inputenc}

\usepackage{microtype}

\usepackage{inconsolata}

\usepackage{graphicx}

\usepackage{hyperref}
\title{Diffusion Models Through a Global Lens: Are They Culturally Inclusive?}

\author{
\textbf{Zahra Bayramli}$^{1,3}$\thanks{\ Equal contribution. Co-first authors: \\ \texttt{\{zahrabayramly, ayhansuleymanzade\}@kaist.ac.kr}}%
\thanks{\ Work done during undergraduate studies at the School of Computing, KAIST.}%
~~ \textbf{Ayhan Suleymanzade}$^{1}$\footnotemark[1]%
~~ \textbf{Na Min An}$^{2}$%
~~ \textbf{Huzama Ahmad}$^{2}$ \\
\textbf{Eunsu Kim}$^{1}$%
~~ \textbf{Junyeong Park}$^{1}$%
~~ \textbf{James Thorne}$^{2}$\thanks{\ Corresponding author}%
~~ \textbf{Alice Oh}$^{1}$\footnotemark[3] \\
$^1$School of Computing, KAIST \quad
$^2$Graduate School of AI, KAIST \quad
$^3$SOCAR \\
}

\begin{document}
\maketitle
\begin{abstract}

Text-to-image diffusion models have recently enabled the creation of visually compelling, detailed images from textual prompts. However, their ability to accurately represent various cultural nuances remains an open question. In our work, we introduce \textsc{CultDiff} benchmark, evaluating whether state-of-the-art diffusion models can generate culturally specific images spanning ten countries. We show that these models often fail to generate cultural artifacts in architecture, clothing, and food, especially for underrepresented country regions, by conducting a fine-grained analysis of different similarity aspects, revealing significant disparities in cultural relevance, description fidelity, and realism compared to real-world reference images. With the collected human evaluations, we develop a neural-based image-image similarity metric, namely, \textsc{CultDiff-S}, to predict human judgment on real and generated images with cultural artifacts. Our work highlights the need for more inclusive generative AI systems and equitable dataset representation over a wide range of cultures.
\end{abstract}

\section{Introduction}
Text-to-image (T2I) diffusion models recently have shown a significant advance in generating high-quality images from text prompts\citep{saharia2022photorealistic, ramesh2022hierarchical, dalle-3, podell2023sdxl, blattmann2023align, esser2024scaling}. Remarkable success has been demonstrated in producing realistic and detailed visual representations, ranging from generating complex scenes \cite{liu2024draw} to performing style transfer using diffusion models \cite{wang2023stylediffusion, zhang2023inversion}. While it is crucial to address the challenge of generating culture-aware imagery, current diffusion models have yet to be evaluated on their ability to create a distribution of accurate and fair representations of images \cite{cao2024exploring,liu2023towards}. 

A recent study, \citet{basu2023inspecting} reveals that while current diffusion models can generate realistic and contextually relevant images, they frequently fail to achieve geographical inclusiveness, generating globally representative images. Furthermore, a critical gap remains in these models' ability to produce culturally accurate representations, failing to create concepts or artifacts for particular cultures, thus, lacking in culture awareness  \cite{kannenbeyond}.
While benchmark datasets have been developed to capture cultural diversity \cite{jha-etal-2023-seegull, myung2024blend, li2024foodieqa}, to the best of our knowledge, there is limited prior work exploring the ability of diffusion models to generate culturally specific artifacts, particularly those representing underrepresented cultures \cite{kannenbeyond, nayak2024benchmarking}.

The primary motivation of our study is to investigate the capability of state-of-the-art diffusion models in accurately representing cultural themes for both overrepresented and underrepresented cultures.  We consider culture as societal constructs tied to country boundaries and use `culture' and `country' interchangeably in this paper, similar to prior approaches \cite{kannenbeyond, li2024culture}. The overrepresented cultures defined here include countries from Western, European, Industrialized, Rich, and Democratic (WEIRD) regions \cite{henrich2010weirdest} (\textit{e.g.}, United States, United Kingdom), as well as countries from Asia, based on their Asia Power Index\footnote{https://power.lowyinstitute.org} (\textit{e.g.}, China, South Korea). In contrast, countries where the primary languages (\textit{e.g.}, Amharic, Azerbaijani) have limited resource availability, as defined in \citet{joshi2020state}, are considered underrepresented (\textit{e.g.}, Ethiopia, Azerbaijan).

In this paper, we present a new benchmark dataset designed to evaluate these models' ability to generate images that accurately reflect both high-resource and low-resource cultures\footnote{We use the terms over/underrepresented and high/low-resource interchangeably throughout the paper.}. Our benchmarking dataset includes a curated collection of prompts for architectural landmarks, clothing, and food from various countries to ensure comprehensive coverage of cultural diversity. Since existing image similarity metrics (\textit{e.g.}, FID \cite{heusel2017gans}) struggle to evaluate cultural nuances accurately, we propose a model-based metric trained and validated with human feedback and compare its correlation. Our contributions can be summarized as follows:
\begin{itemize}
    \item \textbf{CultDiff benchmark dataset}: We introduce a novel dataset, CultDiff, specifically designed to evaluate diffusion models on their ability to generate culturally accurate images across ten countries with varying resource levels, incorporating human evaluations.

    \item \textbf{Analysis on various culture representations of generated images}: We explore how well diffusion models represent low- and high-resource cultures, providing a fine-grained analysis of different aspects of similarity (\textit{e.g.}, image-image, image-description).

   \item \textbf{An automatic image-image similarity evaluation metric}: We present a similarity metric trained and validated with human feedback and compare its capabilities to existing similarity metrics, demonstrating its potential to better capture cultural nuances in generated images.
\end{itemize}

\begin{figure*}[t] 
  \centering
\includegraphics[width=\textwidth,height=0.44\textheight,keepaspectratio]{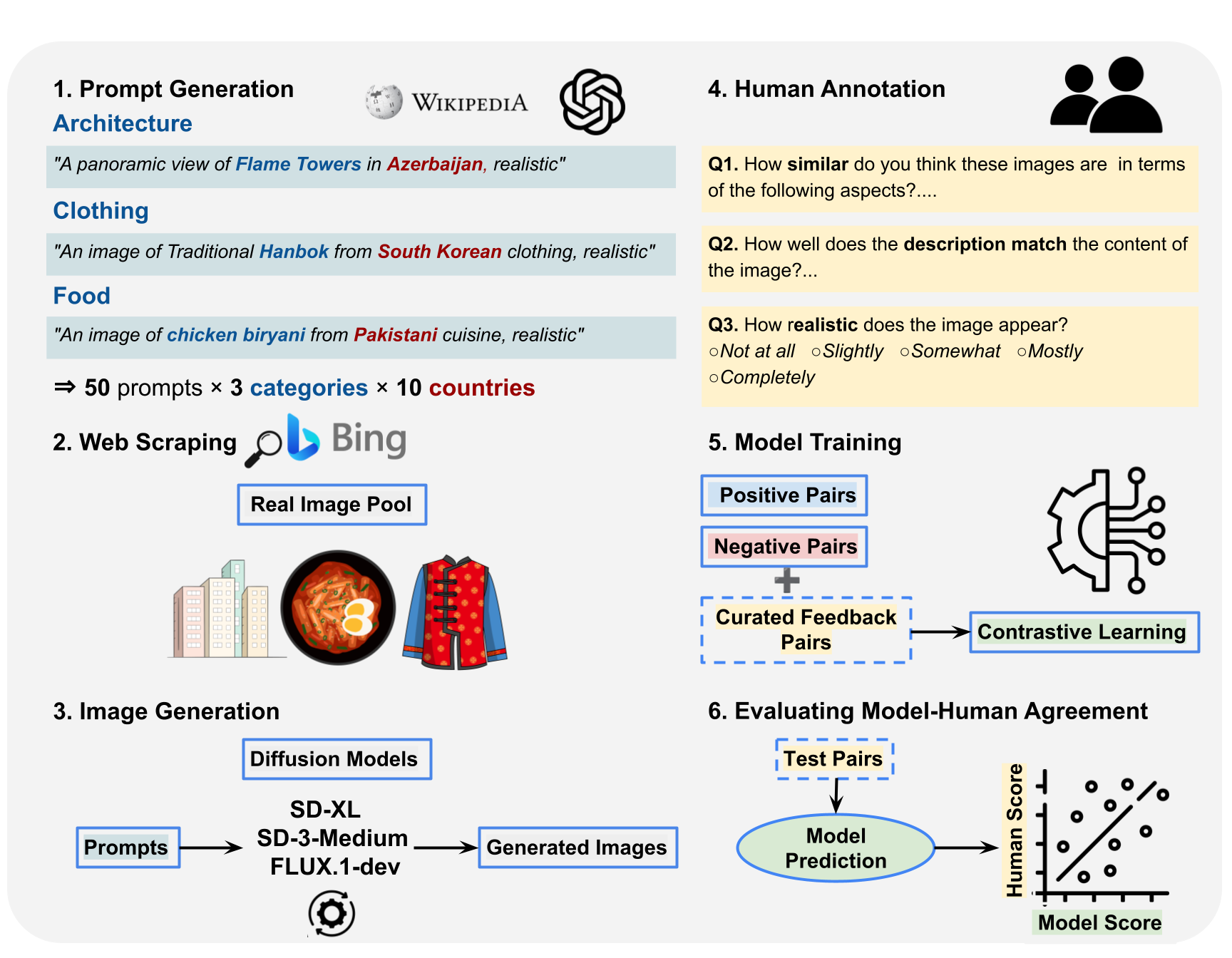}
  \caption{The overall framework for image generation, annotation, and evaluation, consisting of six steps. (1) 50 text prompts for three categories—architecture, food, and clothing—across 10 countries were generated. (2) Reference images were collected using Bing search. (3) Three diffusion models (SD-XL, SD-3-Medium, FLUX.1-dev) were used to generate images. (4) Human annotators evaluated images based on similarity, description match, and realism. (5) Human feedback was used to create positive and negative pairs for contrastive learning. (6) Model performance was assessed by measuring agreement with human scores.}
  \label{fig:framework}
\end{figure*}

\section{Related Works}
\subsection{Cultural-Aware Datasets}
Beginning with the benchmark to evaluate image quality from simpler text prompts using text-to-image (T2I) diffusion models, such as HPDv2  \cite{wu2023human} using human preferences, more recently advanced efforts include creating comprehensive benchmarks such as T2I-CompBench  \cite{huang2023t2i} and GenAI-Bench  \cite{li2024genai}. These prior benchmarks assess the performance of T2I models across various aspects, including realism, fidelity, and compositional text-to-image generation. Although a very recent study \cite{jha2024visage} has started investigating global representation in T2I models, their emphasis has primarily been on regional social stereotypes. From another perspective, \citet{khanuja2024image} proposed the task of image transcreation for cultural relevance demonstrating that image-editing models struggle to adapt images meaningfully across cultures.

On the other hand, many widely-used datasets for training T2I synthesis models, such as LAION-400M  \cite{schuhmann2021laion}, tend to exhibit Anglo-centric and Euro-centric biases, as noted by  \citet{birhane2021multimodal}. These biases skew the representation of cultures in generated images, often favoring Western perspectives \cite{mihalcea2024ai}. In response to this, several researchers have worked to create datasets that better represent diverse cultures. For example, the MaRVL dataset  \cite{liu2021visually} was specifically designed to include a broader array of languages and cultural concepts, covering regions such as Indonesia, Swahili-speaking Africa, Tamil-speaking South Asia, Turkey, and Mandarin-speaking China and addressing biases in datasets that predominantly focused on North American and Western European cultures. Likewise, Culturally-aware Image Captioning (CIC) framework \cite{yun2024cic} was proposed to enhance cultural diversity in image descriptions.

In the Large Language Models (LLMs) domains, SeeGULL dataset  \cite{jha-etal-2023-seegull} broadens stereotype benchmarking to encompass global and regional diversity, and BLEnD benchmark  \cite{myung2024blend} evaluates the cultural knowledge of LLMs across various languages, including low-resource ones.
Similarly, the Dollar Street dataset \cite{gaviria2022dollar} sought to capture everyday life across a wide variety of socioeconomic backgrounds, presenting a more globally inclusive view. Furthermore, CulText2I \cite{ventura2023navigating} was introduced as a multilingual benchmark and evaluation framework to analyze cultural encoding in TTI models across languages, cultural concepts, and domains, highlighting how language impacts cultural representation in generated images. In addition, \citet{liu2023towards} introduced the CCUB dataset, which was developed to promote cultural inclusivity by collecting images and captions representing the cultural contexts of different countries. \citet{liu2024cultural} introduced a Challenging Cross-Cultural (C3) benchmark to evaluate failures in culturally specific generation and proposed fine-tuning on curated data. 
In the multimodal large language model (MLLM) domain, recent benchmarks such as CVQA \cite{romero2024cvqa} focused on culturally aware question answering by capturing nuances across a diverse set of languages. Most recently, CUBE \cite{kannenbeyond} is a large-scale dataset of cultural artifacts spanning 8 countries across different geo-cultural regions for evaluating cultural diversity. Our work further contributes to dataset creation by expanding the focus to include low-resource cultures, thereby addressing gaps between overrepresented and underrepresented cultural contexts.

\subsection{Diffusion Model Evaluation}
Several metrics have been developed and widely used to evaluate the quality of images generated by T2I models. These include measures of realism such as the Inception Score (IS) \citep{salimans2016improved}, Fréchet Inception Distance (FID) \citep{heusel2017gans}, and Image Realism Score (IRS) \citep{chen2023quantifying}. In particular, IS evaluates image quality and diversity based on classification probabilities, FID quantifies the similarity between generated and real image distributions, and IRS mainly analyzes basic image characteristics to measure the realism of the visual content.
In addition, the alignment between generated images and the corresponding prompts has been evaluated using various metrics, such as CLIPScore \citep{hessel2021clipscore}, VQA Score \citep{lin2025evaluating}, and ImageReward \citep{xu2023imagereward}, which incorporates human feedback to enhance T2I models further. Although \citet{kannenbeyond} has explored cultural diversity in text-to-image model generations using the Vendi Score \citep{nguyen2024quality}, measuring cultural accuracy in generated images has not yet been successfully achieved with existing metrics. As a result, the most reliable approach remains relying on human participants, as demonstrated in several works \citep{kannenbeyond, nayak2024benchmarking}. In our work, we also utilize human annotation; however, due to the time and cost associated with it, there is a pressing need for automatic evaluation. To address this, we present a metric specifically trained on a culture-aware dataset to observe its potential for aligning with human preferences.

\section{\textsc{CultDiff} Benchmark}
This section outlines the process of building our \textsc{CultDiff} benchmark, including the data construction (steps 1-3 in Figure~\ref{fig:framework} and Section~\ref{subsec:3.1}), human annotation (step 4 in Figure~\ref{fig:framework} and Section~\ref{subsec:3.2}), and design of the architectural framework for the metric evaluation (steps 5-6 in Figure~\ref{fig:framework} and Section~\ref{subsec:3.3}).

\subsection{\textsc{CultDiff} Dataset Construction}\label{subsec:3.1}
In this section, we present the construction of the   \textsc{CultDiff} dataset, which is designed to evaluate the performance of diffusion models in generating culturally relevant images across various artifact categories and countries. The dataset creation process is composed of three key steps as outlined in Figure~\ref{fig:framework}: (1) prompt generation, (2) collection of real images, and (3) synthetic image generation.

\paragraph{Prompt Generation.}
To create the benchmark dataset for evaluating the diffusion models, we focused on three artifact categories: architecture (or landmarks), clothing, and food across ten countries. 
The prompt generation process involved the following steps: (1)\textbf{Artifact Collection:} Compiled a list of 50 artifact names per category and country using Wikipedia, cultural heritage websites, and travel platforms\footnote{https://www.tripadvisor.com} for landmarks. (2) \textbf{Prompt Engineering:} Created \textit{custom-generated prompts} for each artifact which are simple, structured prompts (e.g., "A panoramic view of \{landmark\} in \{country\}, realistic.") similar to the prompts used in \citet{kannenbeyond}  (refer to Appendix~\ref{sec:appendix_a1} for the list of prompts).

\paragraph{Image Collection and Generation.}
To acquire real images for training and evaluation, we collected images for each artifact across all categories and countries from the internet using the generated prompts. Recognizing that real-world images vary significantly due to factors such as location, camera angle, lighting, and weather conditions, we aimed to minimize potential biases by selecting five images instead of just one for each artifact. The images were scraped using Bulk Bing Image Downloader (BBID)\footnote{https://github.com/ostrolucky/Bulk-Bing-Image-downloader}, an open-source Python tool that enables fast downloads and filters out adult content. 

The image generation process involved synthesizing images based on the generated prompts. We used three diffusion models: Stable Diffusion XL \citep{podell2023sdxl}, Stable Diffusion 3 Medium Diffusers \citep{esser2024scaling}, and FLUX.1-dev\footnote{We refer to FLUX.1-dev as FLUX throughout this paper.} \citep{flux2023}. 

\subsection{Human Evaluation}\label{subsec:3.2}
\begin{figure}[t]
  \includegraphics[width=\columnwidth]{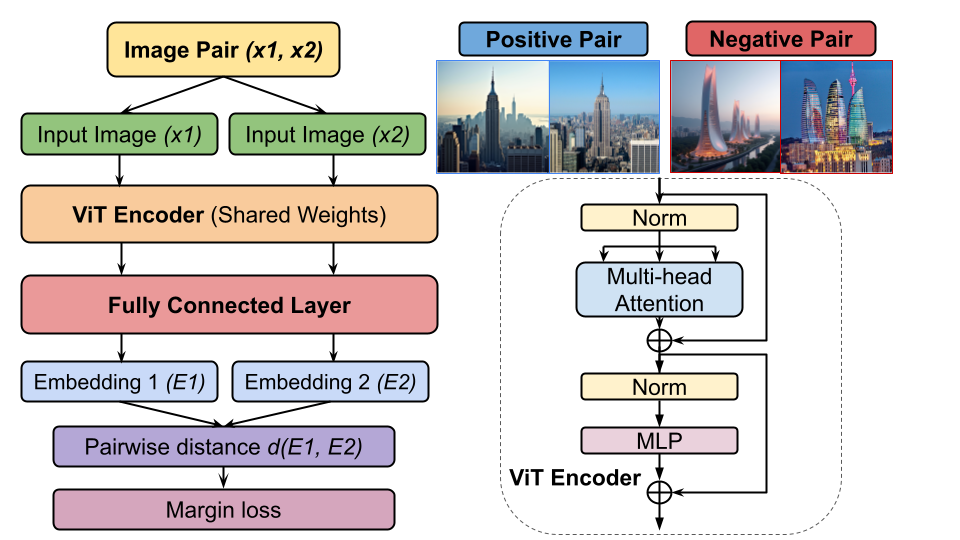}
  \caption{Proposed contrastive learning pipeline with image pairs using Vision Transformer (ViT).}
  \label{fig:vit_architecture}
\end{figure}

A critical part of our experimental design involved human annotation to assess the cultural and contextual relevance of AI-generated images. We first created a survey comprising 150 images for all ten countries (\textit{e.g.}, United States, China, Azerbaijan, and Indonesia), where the images for each country were equally divided across three categories: architecture, clothing, and food items. Each survey included a set of four images for each artifact: three real images (collected via web scraping) and one AI-generated image (produced by one of the three diffusion models).

To prevent model bias, each human annotator was distributed with randomly selected images generated from one of three diffusion models for each question. Hence, the degree of annotator agreement implies varying image quality generated by different models (Appendix~\ref{sec:appendix_a3} for details). Three human annotators from each country participated in the study, ensuring cultural familiarity and context-specific insights. All 30 annotators were asked to evaluate each image based on the following multiple-choice questions:

\paragraph{Image-image similarity scores} are collected for each category as follows:
   \begin{itemize}
       \item \textbf{Architecture:} How similar is the rightmost image (\textit{i.e.}, AI-generated image) to the first three images (\textit{i.e.}, real reference images) in terms of overall similarity, shapes, materials/textures, and background/surroundings? 
       
       \item \textbf{Clothing:} How similar is the rightmost image to the first three images in terms of overall similarity, color/texture, design/patterns, and background elements? 
       
       \item \textbf{Food:} How similar is the rightmost image to the first three images regarding overall similarity, presentation/plating, colors/textures, and ingredients/components?
   \end{itemize}

\paragraph{Image-text alignment scores} are calculated \textit{how well the description} (\textit{i.e.}, prompts mentioned in Section~\ref{subsec:3.1}) \textit{matches the content of the AI-generated image?}

\paragraph{Realism scores} are assessed based on \textit{how realistic the AI-generated image appears}.

Each question for these three multiple-choice subquestions was rated using a Likert scale of 1 to 5: (1) Not at all, (2) Slightly, (3) Somewhat, (4) Mostly, (5) Completely, allowing annotators to express their level of agreement or similarity with the presented image (detailed in Appendix~\ref{sec:appendix_a3}). We recruited 30 annotators from 10 countries: Azerbaijan, Pakistan, Ethiopia, South Korea, Indonesia, China, Spain, Mexico, the United States, and the United Kingdom. Annotators from the first five countries were recruited through college communities, while those from the remaining countries were recruited via the Prolific\footnote{https://www.prolific.com} crowdsourcing platform. The recruitment on Prolific involved screening participants based on their location, nationality, and birthplace.

\subsection{Model Evaluation}\label{subsec:3.3}

\subsubsection{Pair and Dataset Creation}
Our training data consists of real image and real-synthetic image pairs, ensuring a rich, wide variety of training samples. Specifically, we constructed the following image pairs:

\textbf{Real image pairs:} Positive pairs consist of real images of the same artifact (e.g., two images of the Flame Towers in Azerbaijan), while negative pairs include real images of different artifacts from the same country or any artifacts from different countries (e.g., the Flame Towers vs. the Empire State Building).

\textbf{Real-synthetic image pairs:} The positive image pairs generated by diffusion models (as described in Section~\ref{subsec:3.1}) with an average image-image similarity score $\geq 3$, paired with three real images used in the human evaluation. The negative pairs consist of AI-generated images with an average image-image similarity score $< 3$, paired with three real images similar to the positive pairs.

We split our dataset using disjoint prompts for each set to ensure no overlap between the training, validation, and test sets. We selected 30 random prompts for the training set, 10 for the validation set, and 10 for the test set, along with their respective pairs from each model, across all categories and countries. In total, the training set consists of approximately 11k image pairs, while both the validation and test sets contain 2.7k image pairs each. The training set includes both real image pairs and real-synthetic image pairs, while the validation and test sets consist solely of real-synthetic image pairs, with no overlapping images between the sets.
\begin{table*}[ht]
  \centering
  \renewcommand{\arraystretch}{1.2} 
  \setlength{\tabcolsep}{4pt} 
  \resizebox{0.85\textwidth}{!}{ 
    \begin{tabular}{|l|lll|lll|lll|}
      \hline
      \textbf{Diffusion Models} & \multicolumn{3}{c|}{\textbf{Stable Diffusion XL}} & \multicolumn{3}{c|}{\textbf{Stable Diffusion 3 Medium Diffusers}} & \multicolumn{3}{c|}{\textbf{FLUX.1 dev}} \\
      \hline
      \textbf{Category} & \textbf{Architecture} & \textbf{Clothing} & \textbf{Food} & \textbf{Architecture} & \textbf{Clothing} & \textbf{Food} & \textbf{Architecture} & \textbf{Clothing} & \textbf{Food} \\
      \hline

      \textbf{Q1.1-Highest Score} 
      &\cellcolor[HTML]{FFCCCC}\textbf{S. Korea} &\cellcolor[HTML]{ADD8E6}\textbf{Pakistan} &\cellcolor[HTML]{FFCCCC}\textbf{USA} &\cellcolor[HTML]{ADD8E6}\textbf{Azerbaijan} &\cellcolor[HTML]{FFCCCC}\textbf{Mexico} &\cellcolor[HTML]{FFCCCC}\textbf{Mexico} &\cellcolor[HTML]{FFCCCC}\textbf{UK} &\cellcolor[HTML]{FFCCCC}\textbf{Mexico} &\cellcolor[HTML]{FFCCCC}\textbf{USA}\\
     &\cellcolor[HTML]{FFCCCC}UK &\cellcolor[HTML]{FFCCCC}S. Korea &\cellcolor[HTML]{FFCCCC}Spain &\cellcolor[HTML]{FFCCCC}Mexico &\cellcolor[HTML]{FFCCCC}UK &\cellcolor[HTML]{FFCCCC}USA &\cellcolor[HTML]{FFCCCC}S. Korea &\cellcolor[HTML]{ADD8E6}Pakistan &\cellcolor[HTML]{ADD8E6}Pakistan\\
     &\cellcolor[HTML]{FFCCCC}Mexico &\cellcolor[HTML]{FFCCCC}Mexico &\cellcolor[HTML]{ADD8E6}Pakistan &\cellcolor[HTML]{FFCCCC}S. Korea &\cellcolor[HTML]{FFCCCC}S. Korea &\cellcolor[HTML]{ADD8E6}Pakistan &\cellcolor[HTML]{FFCCCC}Mexico &\cellcolor[HTML]{FFCCCC}S. Korea &\cellcolor[HTML]{FFCCCC}Mexico\\   
     &\cellcolor[HTML]{FFCCCC}USA &\cellcolor[HTML]{FFCCCC}USA &\cellcolor[HTML]{FFCCCC}Mexico &\cellcolor[HTML]{FFCCCC}UK &\cellcolor[HTML]{ADD8E6}Pakistan &\cellcolor[HTML]{FFCCCC}Spain &\cellcolor[HTML]{FFCCCC}USA &\cellcolor[HTML]{FFCCCC}UK &\cellcolor[HTML]{FFCCCC}UK\\
    &\cellcolor[HTML]{ADD8E6}Ethiopia &\cellcolor[HTML]{FFCCCC}UK &\cellcolor[HTML]{ADD8E6}Indonesia &\cellcolor[HTML]{ADD8E6}Ethiopia &\cellcolor[HTML]{ADD8E6}Indonesia &\cellcolor[HTML]{FFCCCC}UK &\cellcolor[HTML]{ADD8E6}Ethiopia &\cellcolor[HTML]{FFCCCC}USA &\cellcolor[HTML]{FFCCCC}Spain\\
    \hline
    &\cellcolor[HTML]{FFCCCC}Spain &\cellcolor[HTML]{ADD8E6}Indonesia &\cellcolor[HTML]{FFCCCC}UK &\cellcolor[HTML]{FFCCCC}USA &\cellcolor[HTML]{FFCCCC}USA &\cellcolor[HTML]{ADD8E6}Indonesia &\cellcolor[HTML]{ADD8E6}Azerbaijan &\cellcolor[HTML]{ADD8E6}Indonesia &\cellcolor[HTML]{ADD8E6}Azerbaijan\\
    &\cellcolor[HTML]{ADD8E6}Azerbaijan &\cellcolor[HTML]{ADD8E6}Azerbaijan &\cellcolor[HTML]{ADD8E6}Azerbaijan &\cellcolor[HTML]{ADD8E6}Indonesia &\cellcolor[HTML]{ADD8E6}Azerbaijan &\cellcolor[HTML]{ADD8E6}Azerbaijan &\cellcolor[HTML]{FFCCCC}China &\cellcolor[HTML]{ADD8E6}Azerbaijan &\cellcolor[HTML]{FFCCCC}China\\
    &\cellcolor[HTML]{ADD8E6}Pakistan &\cellcolor[HTML]{FFCCCC}Spain &\cellcolor[HTML]{FFCCCC}S. Korea &\cellcolor[HTML]{FFCCCC}China &\cellcolor[HTML]{FFCCCC}Spain &\cellcolor[HTML]{ADD8E6}Ethiopia &\cellcolor[HTML]{ADD8E6}Indonesia &\cellcolor[HTML]{FFCCCC}Spain &\cellcolor[HTML]{ADD8E6}Indonesia\\
    &\cellcolor[HTML]{ADD8E6}Indonesia &\cellcolor[HTML]{ADD8E6}Ethiopia &\cellcolor[HTML]{ADD8E6}Ethiopia &\cellcolor[HTML]{FFCCCC}Spain &\cellcolor[HTML]{ADD8E6}Ethiopia &\cellcolor[HTML]{FFCCCC}China &\cellcolor[HTML]{ADD8E6}Pakistan &\cellcolor[HTML]{FFCCCC}China &\cellcolor[HTML]{FFCCCC}S. Korea\\
    \textbf{Q1.1-Lowest Score} &\cellcolor[HTML]{FFCCCC}\textbf{China} &\cellcolor[HTML]{FFCCCC}\textbf{China} &\cellcolor[HTML]{FFCCCC}\textbf{China} &\cellcolor[HTML]{ADD8E6}\textbf{Pakistan} &\cellcolor[HTML]{FFCCCC}\textbf{China} &\cellcolor[HTML]{FFCCCC}\textbf{S. Korea} &\cellcolor[HTML]{FFCCCC}\textbf{Spain} &\cellcolor[HTML]{ADD8E6}\textbf{Ethiopia} &\cellcolor[HTML]{ADD8E6}\textbf{Ethiopia} \\
      \hline
      \textbf{Q1.2-Highest Score} 
      &\cellcolor[HTML]{FFCCCC}\textbf{S. Korea} &\cellcolor[HTML]{FFCCCC}\textbf{S. Korea} &\cellcolor[HTML]{FFCCCC}\textbf{USA} &\cellcolor[HTML]{FFCCCC}\textbf{USA} &\cellcolor[HTML]{FFCCCC}\textbf{UK} &\cellcolor[HTML]{FFCCCC}\textbf{USA} &\cellcolor[HTML]{FFCCCC}\textbf{USA} &\cellcolor[HTML]{FFCCCC}\textbf{S. Korea} &\cellcolor[HTML]{FFCCCC}\textbf{USA} \\
      &\cellcolor[HTML]{FFCCCC}Mexico &\cellcolor[HTML]{FFCCCC}UK & \cellcolor[HTML]{FFCCCC}Spain &\cellcolor[HTML]{FFCCCC}Mexico &\cellcolor[HTML]{FFCCCC}S. Korea & \cellcolor[HTML]{FFCCCC}Spain &\cellcolor[HTML]{FFCCCC}UK &\cellcolor[HTML]{ADD8E6}Indonesia &\cellcolor[HTML]{FFCCCC}Spain \\
      &\cellcolor[HTML]{FFCCCC}UK & \cellcolor[HTML]{FFCCCC}Mexico &\cellcolor[HTML]{ADD8E6}Pakistan &\cellcolor[HTML]{FFCCCC}S. Korea & \cellcolor[HTML]{FFCCCC}Mexico &\cellcolor[HTML]{ADD8E6}Pakistan & \cellcolor[HTML]{FFCCCC}Mexico &\cellcolor[HTML]{FFCCCC}Mexico &\cellcolor[HTML]{ADD8E6}Azerbaijan \\
      &\cellcolor[HTML]{FFCCCC}USA &\cellcolor[HTML]{FFCCCC}USA &\cellcolor[HTML]{ADD8E6}Indonesia &\cellcolor[HTML]{ADD8E6}Azerbaijan &\cellcolor[HTML]{ADD8E6}Indonesia &\cellcolor[HTML]{FFCCCC}Mexico &\cellcolor[HTML]{FFCCCC}S. Korea &\cellcolor[HTML]{ADD8E6}Pakistan &\cellcolor[HTML]{ADD8E6}Pakistan \\
      &\cellcolor[HTML]{FFCCCC}Spain &\cellcolor[HTML]{ADD8E6}Indonesia &\cellcolor[HTML]{ADD8E6}Azerbaijan &\cellcolor[HTML]{FFCCCC}UK &\cellcolor[HTML]{ADD8E6}Pakistan &\cellcolor[HTML]{FFCCCC}China &\cellcolor[HTML]{ADD8E6}Ethiopia &\cellcolor[HTML]{FFCCCC}USA & \cellcolor[HTML]{FFCCCC}Mexico \\
      \hline

      &\cellcolor[HTML]{ADD8E6}Ethiopia &\cellcolor[HTML]{ADD8E6}Pakistan &\cellcolor[HTML]{FFCCCC}S. Korea &\cellcolor[HTML]{ADD8E6}Ethiopia &\cellcolor[HTML]{FFCCCC}USA &\cellcolor[HTML]{ADD8E6}Indonesia &\cellcolor[HTML]{ADD8E6}Azerbaijan &\cellcolor[HTML]{FFCCCC}UK & \cellcolor[HTML]{FFCCCC}UK \\
      &\cellcolor[HTML]{ADD8E6}Azerbaijan &\cellcolor[HTML]{ADD8E6}Azerbaijan &\cellcolor[HTML]{FFCCCC}Mexico &\cellcolor[HTML]{ADD8E6}Pakistan &\cellcolor[HTML]{ADD8E6}Azerbaijan &\cellcolor[HTML]{ADD8E6}Azerbaijan &\cellcolor[HTML]{FFCCCC}China &\cellcolor[HTML]{ADD8E6}Azerbaijan &\cellcolor[HTML]{FFCCCC}China \\
      &\cellcolor[HTML]{ADD8E6}Pakistan &\cellcolor[HTML]{FFCCCC}Spain &\cellcolor[HTML]{FFCCCC}UK &\cellcolor[HTML]{FFCCCC}China &\cellcolor[HTML]{FFCCCC}China &\cellcolor[HTML]{FFCCCC}UK &\cellcolor[HTML]{ADD8E6}Pakistan &\cellcolor[HTML]{FFCCCC}Spain &\cellcolor[HTML]{FFCCCC}S. Korea \\
      &\cellcolor[HTML]{FFCCCC}China &\cellcolor[HTML]{ADD8E6}Ethiopia &\cellcolor[HTML]{ADD8E6}Ethiopia &\cellcolor[HTML]{FFCCCC}Spain &\cellcolor[HTML]{ADD8E6}Ethiopia &\cellcolor[HTML]{FFCCCC}S. Korea &\cellcolor[HTML]{FFCCCC}Spain &\cellcolor[HTML]{FFCCCC}China &\cellcolor[HTML]{ADD8E6}Indonesia \\
      \textbf{Q1.2-Lowest Score} &\cellcolor[HTML]{ADD8E6}\textbf{Indonesia} &\cellcolor[HTML]{FFCCCC}\textbf{China} &\cellcolor[HTML]{FFCCCC}\textbf{China} &\cellcolor[HTML]{ADD8E6}\textbf{Indonesia} &\cellcolor[HTML]{FFCCCC}\textbf{Spain} &\cellcolor[HTML]{ADD8E6}\textbf{Ethiopia} &\cellcolor[HTML]{ADD8E6}\textbf{Indonesia} &\cellcolor[HTML]{ADD8E6}\textbf{Ethiopia} &\cellcolor[HTML]{ADD8E6}\textbf{Ethiopia} \\
      \hline
    \end{tabular}
  }
  \caption{\label{table-scores} Ranking of countries by average Q1.1 scores (top half) and Q1.2 scores (bottom half), organized from highest to lowest. The table presents scores across three models for the categories of Architecture, Clothing, and Food. Overrepresented countries are highlighted in red, while underrepresented countries are highlighted in blue.}
\end{table*}

\subsubsection{Similarity Metric}
To build a model-based image-image similarity metric, \textsc{CultDiff-S}, we adopted a contrastive learning approach, utilizing Vision Transformers (ViT) \citep{alexey2020image} to learn embeddings for both real and AI-generated images, as illustrated in Figure~\ref{fig:vit_architecture}. The global attention mechanism in ViT enables effective "local-to-global" correspondence, enhancing self-supervised learning \citep{caron2021emerging, mo2023multi}. This ViT property is particularly important in our work, where it helps evaluate the cultural and contextual relevance of images by capturing both local features and their broader context. To achieve this, we employed a weighted margin loss during the training process, utilizing the scores (collected from our human annotation process explained in Section~\ref{subsec:3.2}) as weights ($w_i$, $i=1, 2, ..., N$ in Eq.~\ref{eq:loss}), motivated by previous studies that utilized GPT feedback scores as the weights in building sentence embedding models \cite{cheng-etal-2023-improving, an2024capturing}. These scores are normalized to 0 to 1 and applied to their corresponding real-synthetic (human-annotated) positive and negative image pairs. For real-image pairs that were not annotated by humans, we assigned a default weight of $w=1$. Our margin loss is defined as:

\begingroup
\small
\[
\mathcal{L} = \frac{1}{N} \sum_{i=1}^{N}  
    w_i \cdot \Big( y_i \cdot d_i^2 + 
    (1 - y_i) \cdot \max(0, m - d_i)^2 
\Big),
\]
\label{eq:loss}
\endgroup
where \( w_i \) denotes the weight derived from the normalized human score when available (default is 1.0), \( d_i \) represents the Euclidean distance between the embeddings of a pair of images (\( d_i = \| f(x_{i1}) - f(x_{i2}) \|_2 \)), and \( f(x) \) is the embedding function learned by the Vision Transformer. The binary label \( y_i \in \{0, 1\} \) indicates whether a pair is positive (\( y_i = 1 \)) or negative (\( y_i = 0 \)). The margin \( m \) enforces a minimum distance between embeddings of negative pairs, while \( N \) is the total number of image pairs in the batch.

This loss function is optimized to make the model learn to (1) push the positive image pairs together and pull the negative image pairs apart and (2) assign importance to pairs with human judgment scores. The second goal guides the model to attend more to the image pairs with higher human scores and less with lower human scores.

\section{Results}
\subsection{Benchmark Results and Analysis}
\subsubsection{Overall Image-Image Similarity}

\begin{figure*}[t]
  \centering
  \includegraphics[width=0.32\textwidth]{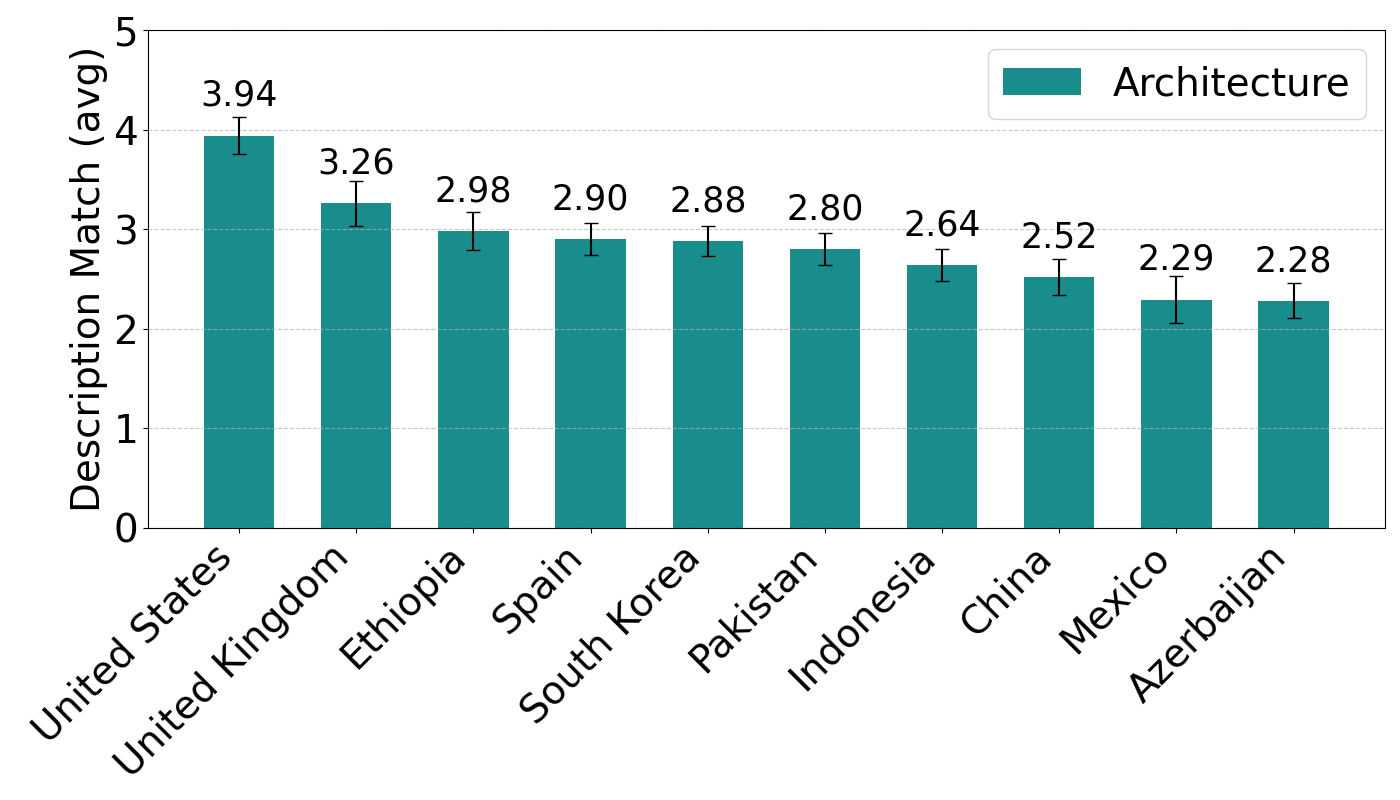}
  \includegraphics[width=0.32\textwidth]{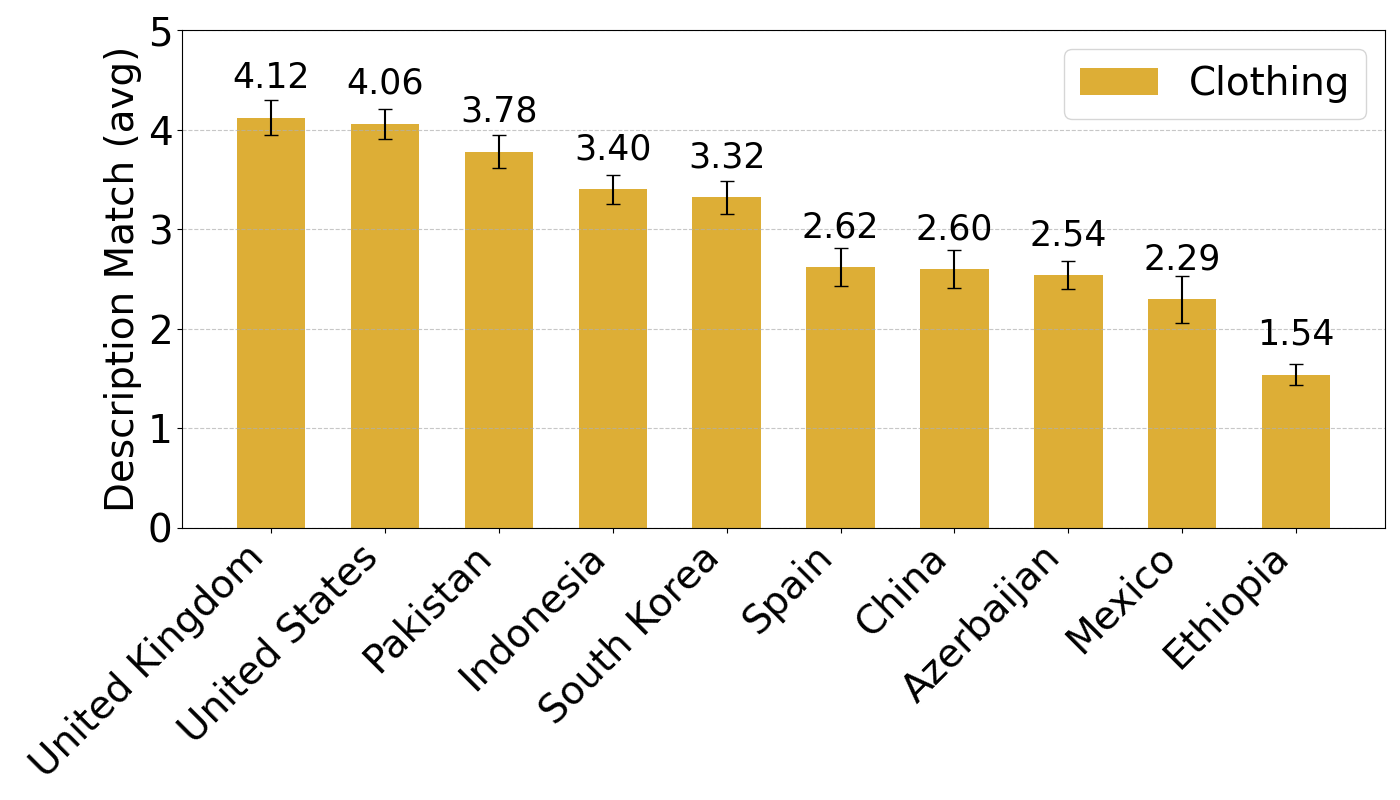}
  \includegraphics[width=0.32\textwidth]{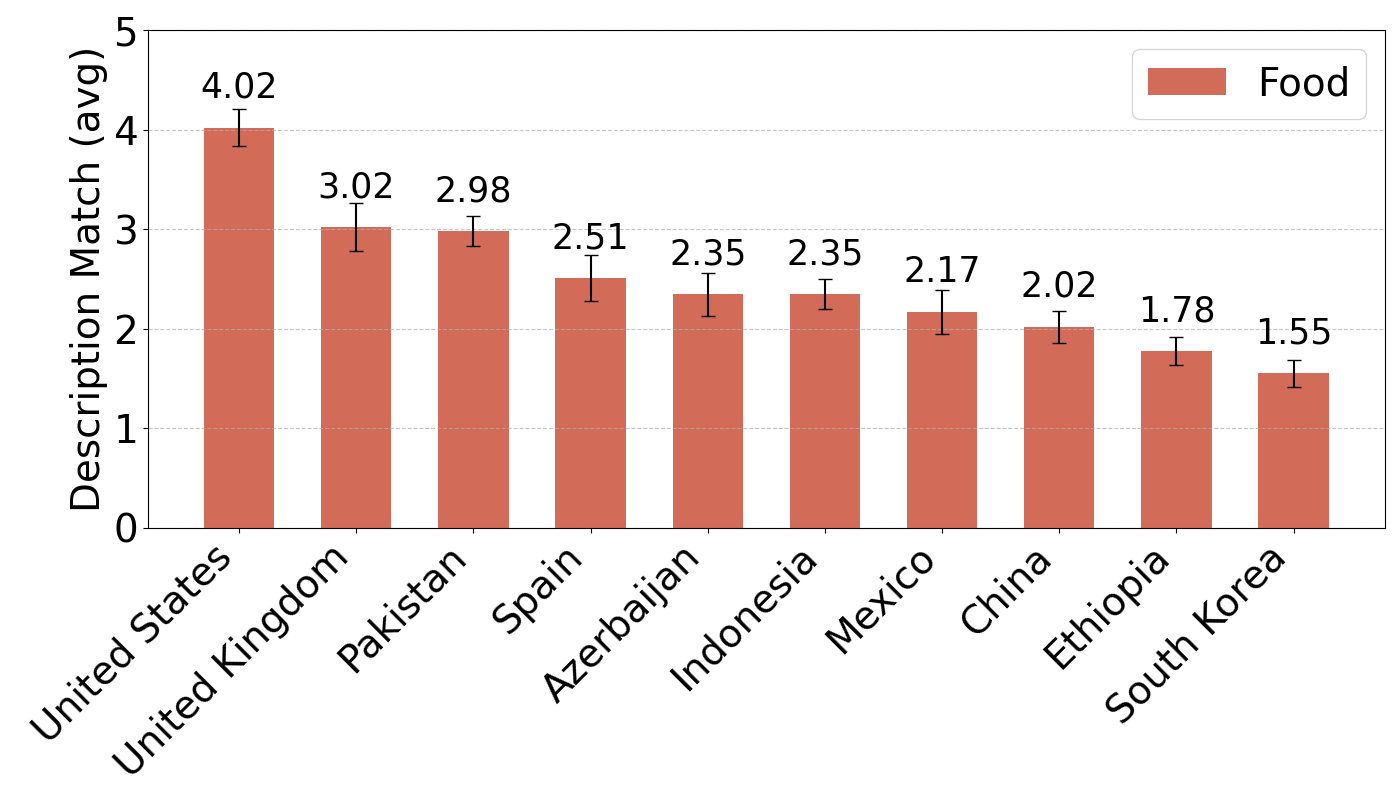}

  \caption{Average description match scores for the FLUX model's outputs across three categories: Architecture (left), Clothing (center), and Food (right). Each bar plot represents the average description match scores for countries within the respective category, ranked from highest to lowest. Error bars indicate the
standard error of the mean.}
  \label{fig:description_match_flux}
\end{figure*}

\begin{figure*}[t]
  \centering
  \includegraphics[width=1.0\textwidth]{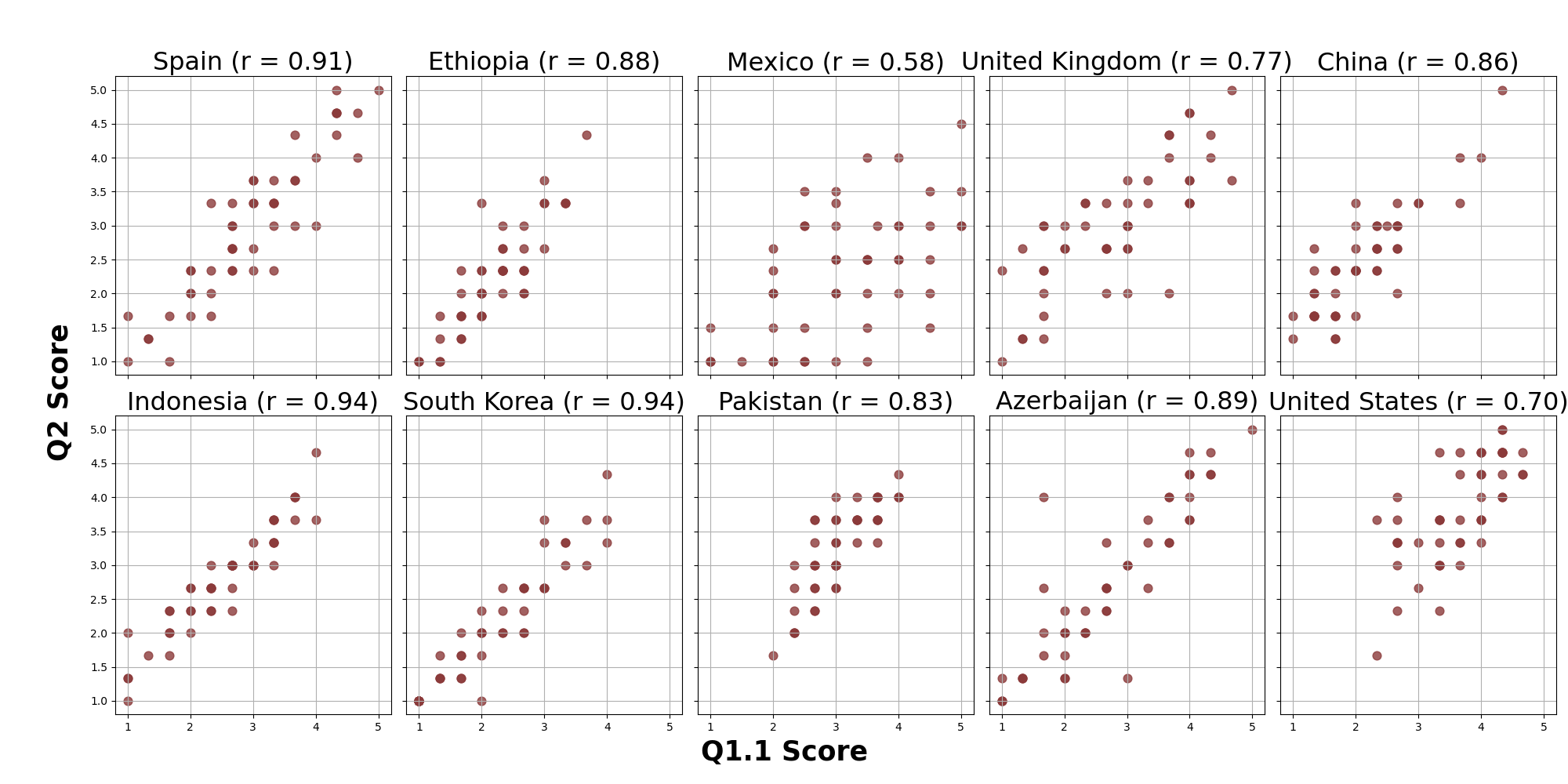}
  \caption{Scatter plots of Q1.1 (horizontal axis) vs. Q2 (vertical axis) scores for Food images, categorized by country. Each subplot displays the country name and corresponding Pearson correlation coefficient $r$.}
  \label{fig:corr_q11_q2_food}
\end{figure*}
We first analyze the similarity results (image-image similarity scores) of the real and generated images from the survey's first question (Q1), which examined various aspects of similarity.
In particular, Q1.1 asked participants to rate the overall similarity. Table~\ref{table-scores} highlights the average scores for all countries across three models. The first half of Table~\ref{table-scores} summarizes the country results: overrepresented countries (South Korea, USA, UK, Mexico) predominantly occupy the top half (highest scores), while underrepresented countries (Azerbaijan, Ethiopia) are mostly in the lower half, with some exceptions.

Q1.2 specifically addressed distinct features: shapes for architectural structures, color/texture for clothing, and presentation/plating for food. The results, summarized in Table~\ref{table-scores}, highlight that the USA, UK, and South Korea consistently ranked among the highest-scoring countries across all categories and models. In contrast, Ethiopia and Indonesia frequently appeared in the lower half of the rankings. Interestingly, China also consistently received some of the lowest scores in this analysis. These findings suggest a need for the development of diffusion models that better represent underrepresented countries, such as Ethiopia and Indonesia, to improve their cultural and contextual accuracy compared to overrepresented countries like the USA, UK, and South Korea.

\subsubsection{Image-Description Match}\label{sec:img_des}

Our analysis focused on the responses to the second question (Q2), which assessed the fidelity of the generated content to the provided description. Figure~\ref{fig:description_match_flux} presents the results for the FLUX model's outputs across architecture, clothing, and food, respectively. The United States and the United Kingdom consistently achieved the highest average fidelity scores across all three categories. In contrast, Azerbaijan, South Korea, and Ethiopia recorded the lowest average scores of 2.28, 1.55, and 1.54 in the architecture, food, and clothing categories, respectively. It is important to note that there is no significant statistical difference between the overrepresented and underrepresented countries in our study, likely due to the small dataset. However, the country rankings remain relevant. 
In fact, these average scores reveal a noticeable gap between the scores of the United States and those of other countries across all categories. These results suggest a potential bias in the model, favoring countries with a more extensive cultural presence on the internet. Addressing this limitation by incorporating more diverse and balanced datasets could improve the model’s fidelity for underrepresented countries and enhance its overall cultural accuracy \cite{ananthram2024see, liu2025culturevlm}. 

Results for Stable Diffusion XL and Stable Diffusion 3 Medium Diffusers (see Appendix~\ref{sec:appendix_a4} for corresponding graphs) also indicate that the United States consistently receives higher scores. Conversely, countries like Azerbaijan and Ethiopia frequently appear in the lower half of the rankings. However, the relative ordering of countries varies across models. 

We also visualized the correlations between responses to the overall similarity question (Q1.1) for real and generated images and the text-image alignment score (Q2). As shown in Figure~\ref{fig:corr_q11_q2_food}, most countries exhibit a strong positive Pearson correlation coefficient ($r$) in the Food category (refer to Appendix~\ref{sec:appendix_a5} for results on architecture and clothing), indicating consistent agreement between Q1.1 and Q2 responses across most countries.

\begin{figure*}[t]
  \centering
  \includegraphics[width=0.32\textwidth]{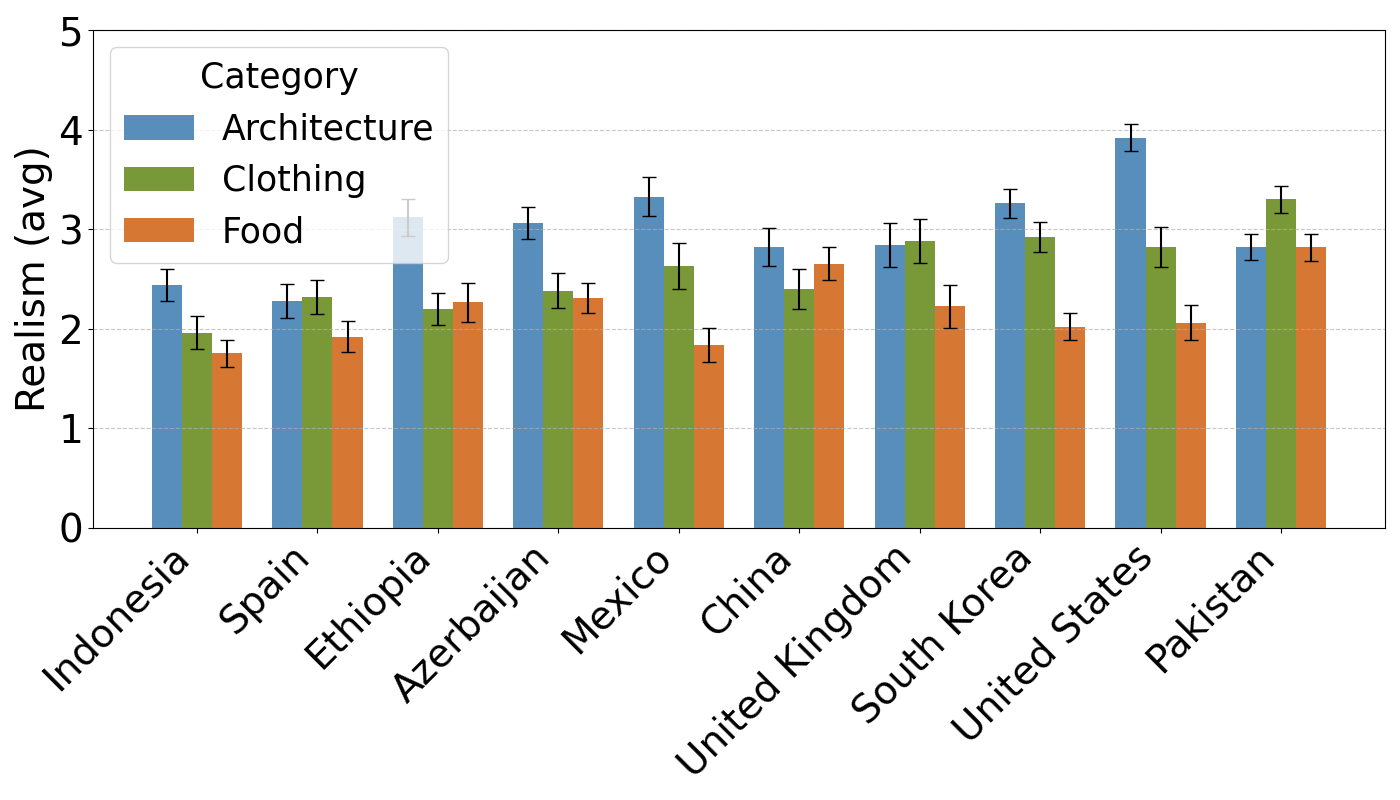}
  \includegraphics[width=0.32\textwidth]{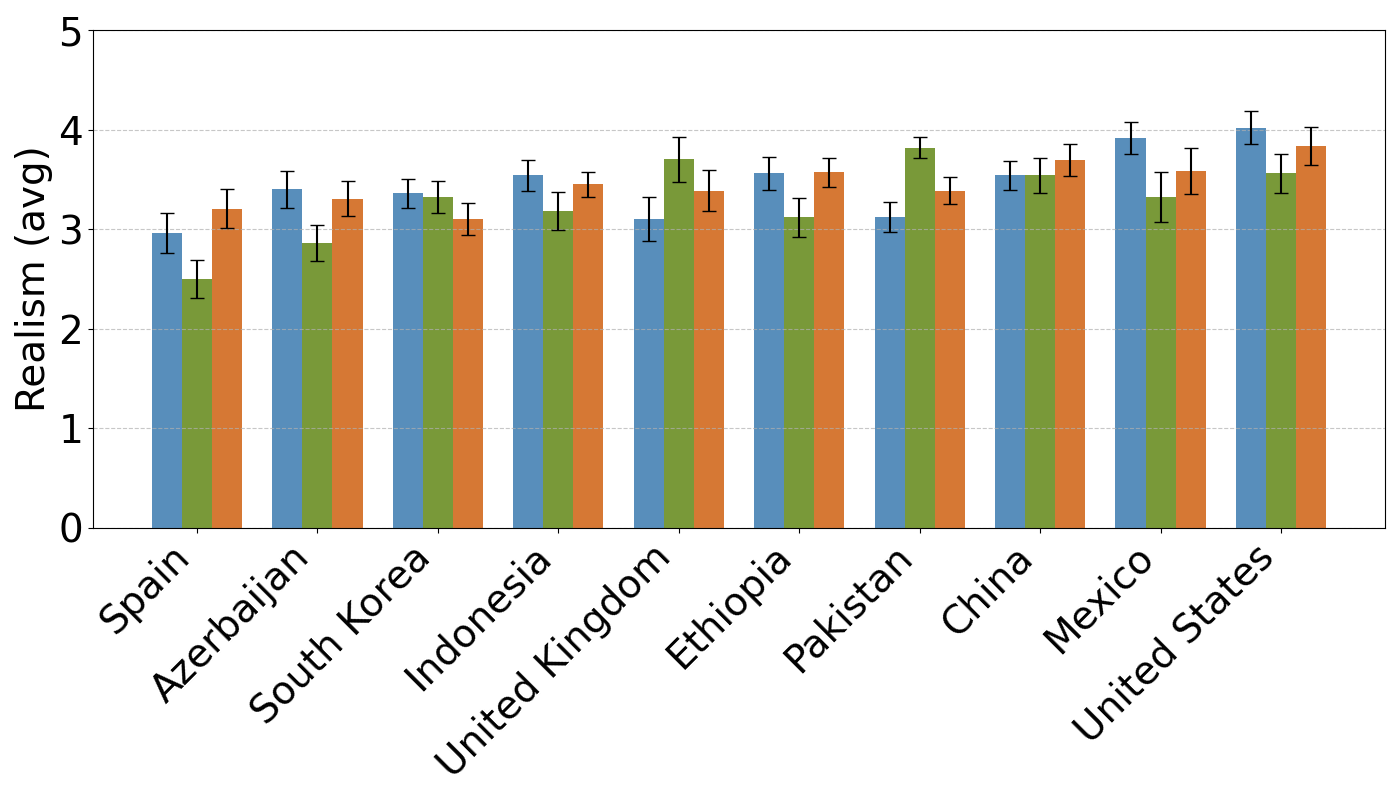}
  \includegraphics[width=0.32\textwidth]{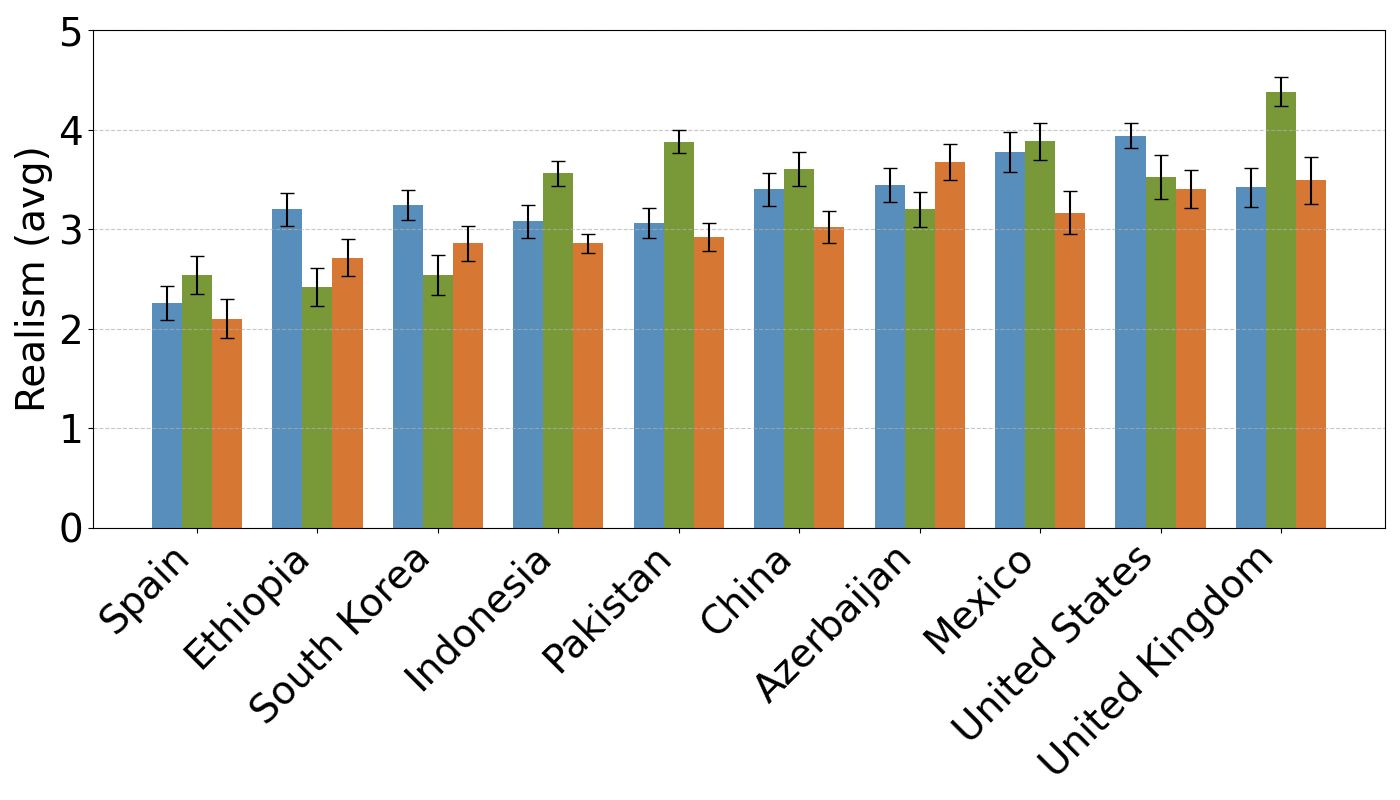}

  \caption{Average realism scores across three models: Stable Diffusion XL (left), Stable Diffusion 3 Medium Diffusers (center), and FLUX (right). The bar plots show realism scores for three categories—Architecture, Clothing, and Food—averaged across countries for each model. Error bars indicate the standard error of the mean.}
  \label{fig:realism_scores}
\end{figure*}
\subsubsection{Realism}\label{sec:realism}
The third question, which focused on rating the realism of the generated images, is visualized in the bar graphs in Figure~\ref{fig:realism_scores}. The FLUX model's output shows that realism scores are relatively higher for the United Kingdom, United States, and Mexico, while Spain, Ethiopia, and South Korea have lower scores. In contrast, the Stable Diffusion 3 Medium model exhibits relatively consistent realism scores across all countries. Meanwhile, the Stable Diffusion XL model demonstrates slightly lower average realism scores across all countries and categories compared to the other two models.

\subsection{Similarity Metric Analysis}
After training the similarity model, we evaluated its performance using our test dataset pairs. For each evaluation pair, we computed additional similarity metrics, including Fréchet Inception Distance (FID) \citep{heusel2017gans}, Learned Perceptual Image Patch Similarity (LPIPS) \citep{zhang2018unreasonable}, and the Structural Similarity Index Measure (SSIM) \citep{wang2004image}. Additionally, we calculated the correlation between image-image similarity scores—derived by averaging the responses to the first four questions in the survey—and each of these similarity metrics. In our metric model, we compute the cosine similarity of the embeddings produced by the model during evaluation as the final similarity score. The results of these correlations are summarized in Table~\ref{tab:correlation_results}. The proposed metric, \textsc{CultDiff-S}, achieves notably higher correlation values, suggesting the usefulness of using our automatic pre-trained image-image metric, showing potential to replace the previous metrics. However, we leave future work to further enhance model correlations with human judgments.
\begin{table}[t!]
\begin{center}
\resizebox{\columnwidth}{!}{%
\begin{tabular}{l|c|c|c|c}
\toprule
\textbf{\makecell{Similarity \\ Metrics}} & \textbf{\makecell{Spearman \\ $\rho$}} & \textbf{\makecell{Pearson \\ $r$}} & \textbf{\makecell{Kendall \\ $\tau_b$}} & \textbf{\makecell{Kendall \\ $\tau_c$}} \\
\midrule
SISM  &  0.0458 &  0.0464 &  0.0315 & 
 0.0324 \\
FID   & -0.1078 & -0.1085 & -0.0768 & -0.0783 \\
LPIPS & -0.1352 & -0.1376 & -0.1002 & -0.1029 \\
\textbf{\textsc{CultDiff-S}} & \textbf{0.1848} &  \textbf{0.1559} & 
 \textbf{0.1278} &  \textbf{0.1312} \\
\bottomrule
\end{tabular}
}
\end{center}
\caption{Correlation between human judgment scores and similarity values produced using different metrics.}
\label{tab:correlation_results}
\end{table}
\section{Discussion}
\subsection{Human-Aligned Image-Image Similarity Evaluation Metric}
Designing an image-image similarity metric that closely correlates with human rankings is inherently difficult due to the complexity of cultural attributes and subjective interpretations. Culture itself is a broad and multifaceted concept, making it challenging to define in a universally accepted manner \cite{liu2024culturally}. In fact, many previous studies in related domains that incorporate cultural aspects either provide only a high-level definition or avoid in-depth discussions on its nuances \cite{adilazuarda2024towards, huang2023culturally, zhou-etal-2023-cultural, cao2024cultural, wan2023personalized}. In our study, we treated countries as representatives of distinct cultures. However, individuals, even within the same country, may have diverse interpretations of cultural artifacts, reflecting regional, ethnic, and personal differences in cultural perception, which can lead to subjective similarity scores given by human annotators.
Additionally, cultural artifacts often evolve over time, influenced by various internal or external factors, further complicating the task of cultural representations. As a result, an effective similarity metric that correlates more closely with human perception must account for both objective attributes, such as structural and visual features, and subjective factors, including historical context, regional variations, and personal interpretations. We believe that developing such a metric necessitates integrating human feedback into the process, combining both data-driven approaches and human-centered evaluations to bridge the gap, as has been demonstrated in other contexts and domains in prior works \cite{xu2023imagereward, stiennon2020learning, ouyang2022training}, as well as in our research.

\subsection{Culturally Aware Diffusion Models}
Recent advancements in fine-tuning diffusion models have demonstrated significant improvement \cite{ruiz2023dreambooth}. Following similar approaches in cultural contexts, fine-tuning diffusion models on smaller, culturally-curated datasets has been shown to lead to more culturally relevant image generation \cite{liu2023towards}. While approaches relying on large culture-specific datasets, such as Japanese Stable Diffusion \cite{japanese_stable_diffusion}, have been generally successful in enhancing cultural awareness for specific cultures, the limited resource presence of underrepresented cultures creates a challenge for applying similar methods. In this context, our dataset further provides a foundation for incorporating data from underrepresented cultures/countries, broadening the reach of culturally inclusive image generation. 
We hope that our work will inspire further research into improving the fairness of T2I generation models, particularly by expanding representation for low-resource cultures in both model training and evaluation.

\section{Conclusion}
In this paper, we assessed the cultural awareness diffusion models spanning 10 countries and 3 categoriessingby our proposed \textsc{CultDiff} Benchmark Dataset. Our human evaluations of various aspects—such as the similarity of AI-generated images to real counterparts, description fidelity, and realism—revealed that the diffusion models exhibit better cultural awareness for high-resource countries than low-resource countries, such as Ethiopia, Azerbaijan, and Indonesia. Furthermore, we proposed an automatic image-image similarity metric that shows improvement in correlating with human judgments to evaluate cultural aspects.

\section*{Limitations}
The main limitation of our work lies in recruiting only three annotators per country due to budget constraints, similar to most human experiments in other studies. While we could derive valuable insights from the human evaluation results, increasing the number of survey participants per country would enhance the robustness and generalizability of our findings. 

\section*{Ethical Statement}
This study was conducted with prior approval from the Institutional Review Board (IRB), ensuring adherence to ethical research standards. All participants recruited through Prolific were compensated at a competitive rate (18 pounds $\sim$ 22.38 USD in January 2025), meeting or exceeding ethical guidelines. Similarly, participants recruited directly, primarily from low-resource countries, were fairly compensated (30K KRW $\sim$ 20.89 USD in January 2025).

To ensure the appropriateness of the materials, all generated images were reviewed by the authors before being provided to the annotators. Additionally, we included a question at the end of the survey asking whether each image was inappropriate or disturbing. All annotators responded "No" to this question, confirming the suitability of the content used in the study.

\section*{Acknowledgments}
This research was supported by the MSIT (Ministry of Science, ICT), Korea, under the National Program for Excellence in SW), supervised by the IITP (Institute of Information \& communications Technology Planing \& Evaluation) in 2024 (2022-0-01092) and National Research Foundation of Korea (NRF) grant funded by the Korea government (MSIT) (No.RS-2024-00406715), and Institute of Information \& communications Technology Planning \& Evaluation (IITP) grant funded by the Korea government (MSIT) (No. RS-2024-00509258 and No. RS-2024-00469482, Global AI Frontier Lab).

This research project has benefited from the Microsoft Accelerate Foundation Models Research (AFMR) grant program through which leading foundation models hosted by Microsoft Azure along with access to Azure credits were provided to conduct the research.

\bibliography{custom}

\clearpage
\newpage

\appendix
\section{Appendix}
\subsection{Prompt Details} \label{sec:appendix_a1}

For each category, we used prompts in the following format: \\ 
1. ``A panoramic view of \{landmark\} in \{country\}, realistic'' \\
2. ``An image of \{clothes\} from \{country\} clothing, realistic''  \\
3. ``An image of \{food\} from \{country\} cuisine, realistic''\\  
For example:  \\
1. ``A panoramic view of the Empire State Building in the United States, realistic''  \\
2. ``An image of Hanfu from Chinese clothing, realistic''  \\
3. ``An image of plov from Azerbaijani cuisine, realistic''  

Additionally, we experimented with various prompting techniques, including GPT-4-generated detailed prompts. However, we observed that these prompts occasionally introduced hallucinations leading to inaccurate image generation. Through empirical analysis, we found that simpler prompts tended to provide more accurate and culturally relevant generations.  

\subsection{Model Parameters} \label{sec:appendix_a2}
\subsubsection{Image Generation}  
The models used for image generation have the following parameter counts: Stable Diffusion XL \cite{podell2023sdxl} has 2.6 billion U-Net parameters, Stable Diffusion 3 Medium \cite{esser2024scaling} has 2 billion parameters, and FLUX.1-dev \cite{flux2023} is a 12-billion-parameter rectified flow transformer.  

\subsubsection{Model Training}  
For model training, we used ViT-Base \cite{alexey2020image}, which has 86 million parameters, 12 layers, a hidden size of 768, an MLP size of 3072, and 12 attention heads. We trained our contrastive learning model using ViT-Base for 10 epochs with a learning rate of $1 \times 10^{-4}$, a batch size of 32, and an image resolution of $224 \times 224$.  

\subsection{Survey Question Details} \label{sec:appendix_a3}

\begin{figure*}[t]
    \centering
    \begin{subfigure}{0.48\textwidth} 
        \includegraphics[width=\textwidth, height=0.5\textheight]{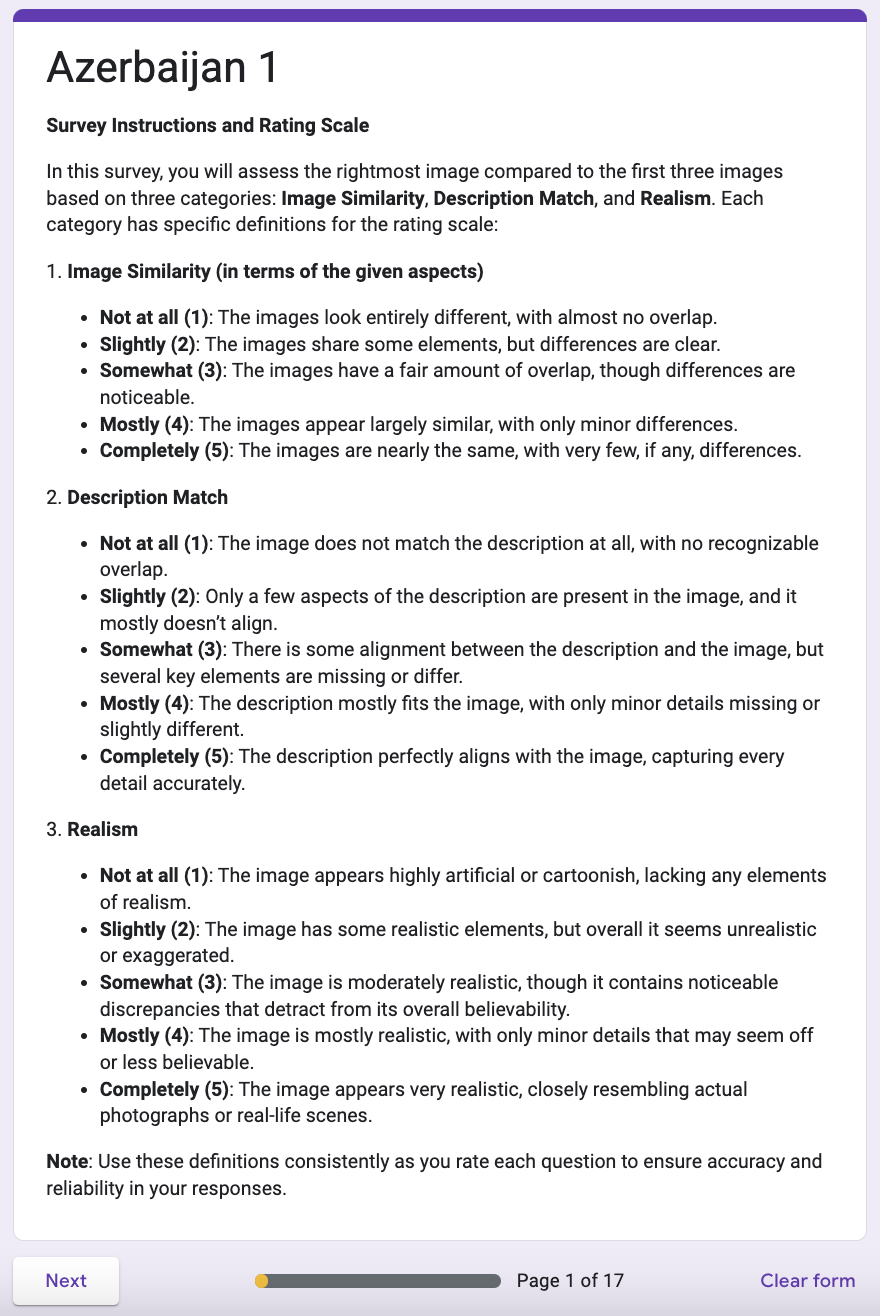}
        \label{fig:survey_instr}
    \end{subfigure}
    \hfill
    \begin{subfigure}{0.48\textwidth} %
        \centering
        \includegraphics[width=0.9\textwidth,height=0.32\textheight]{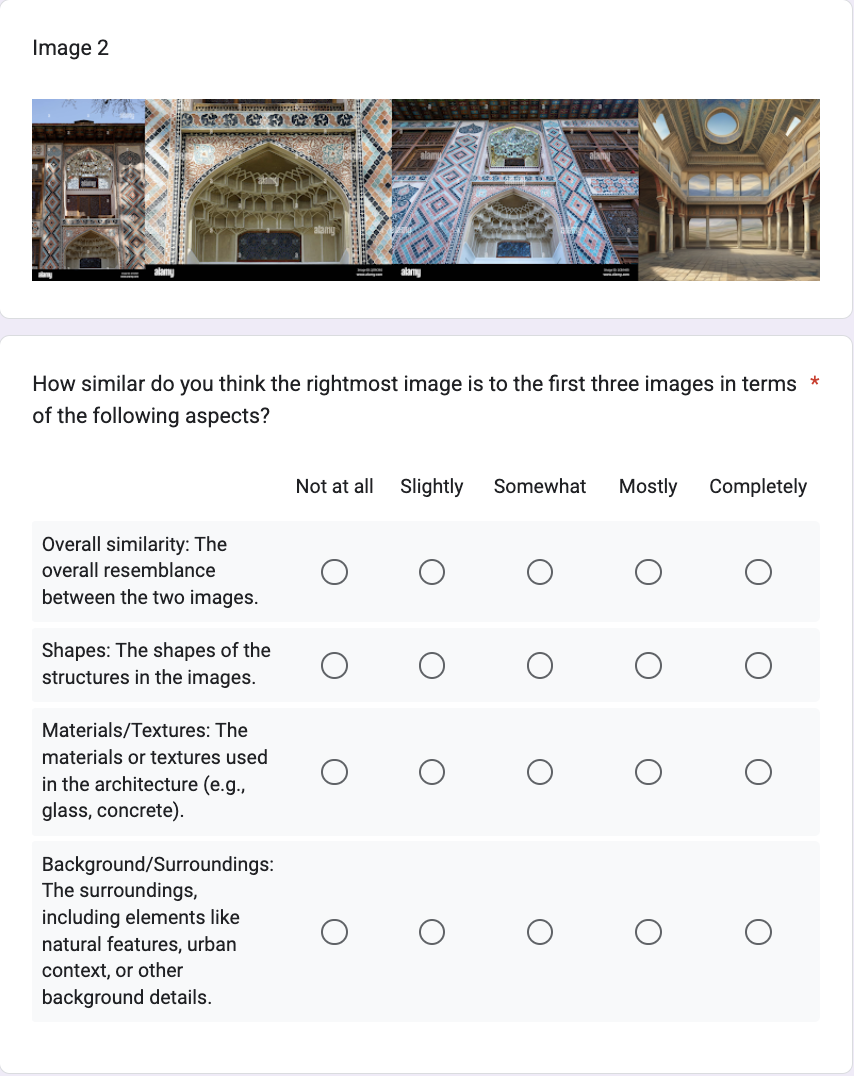}
        \label{fig:survey_q1}
        
        \vspace{0.3em} 
        \includegraphics[width=0.9\textwidth,height=0.18\textheight]{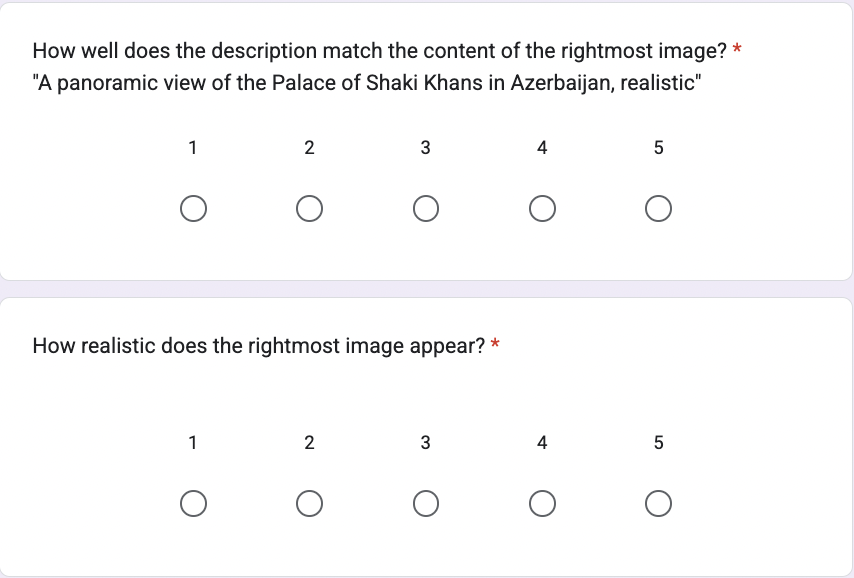}
        \label{fig:survey_q2}
    \end{subfigure}
    \caption{Survey setup illustrating instructions, an image comparison task, and rating questions. The left image shows the survey instructions outlining the tasks and evaluation criteria, while the right side presents examples of the survey questions.}
    \label{fig:survey_overview}
\end{figure*}

This section provides additional details regarding the survey questions and instructions used during the study. Annotators received clear guidance to ensure consistent evaluations of both real and AI-generated images (Figure~\ref{fig:survey_overview} left). Examples of specific survey questions for cultural contexts, such as those related to Azerbaijan, are shown in Figure~\ref{fig:survey_overview} right top and right bottom. The estimated time to complete the survey was approximately 1.5 to 2 hours. Table~\ref{tab:kappa_agree} presents the Fleiss' Kappa scores, reflecting annotator agreement across different diffusion models (Stable Diffusion XL, Stable Diffusion 3 Medium Diffusers, FLUX). The relatively low agreement scores may stem from the fact that annotators from the same country evaluated images generated by different models rather than assessing a shared set of images. Note that we showed a comparative analysis across three models in terms of image-description match (Section~\ref{sec:img_des}) and realism (Section~\ref{sec:realism}).

\begin{table}[ht]
\begin{center}
\resizebox{0.6\columnwidth}{!}{%
\begin{tabular}{l|c}
\toprule
\textbf{Country} & \textbf{\makecell{Fleiss' Kappa \\ $\kappa$}}\\
\midrule
Azerbaijan & 0.14\\
China & 0.13\\
Ethiopia & 0.13\\
Indonesia  &  0.10\\
Mexico & 0.10\\
Pakistan & 0.11\\
South Korea & 0.17\\
Spain & 0.11\\
United Kingdom & 0.12\\
United States & 0.07\\
\bottomrule
\end{tabular}
}
\end{center}
\caption{Inter-annotator agreement across diffusion models, measured by Fleiss' Kappa, for each country.}
\label{tab:kappa_agree}
\end{table}

\subsection{Survey Responses}
\label{sec:appendix_a4}

\begin{figure*}[t]
  \centering
  \includegraphics[width=0.32\textwidth]{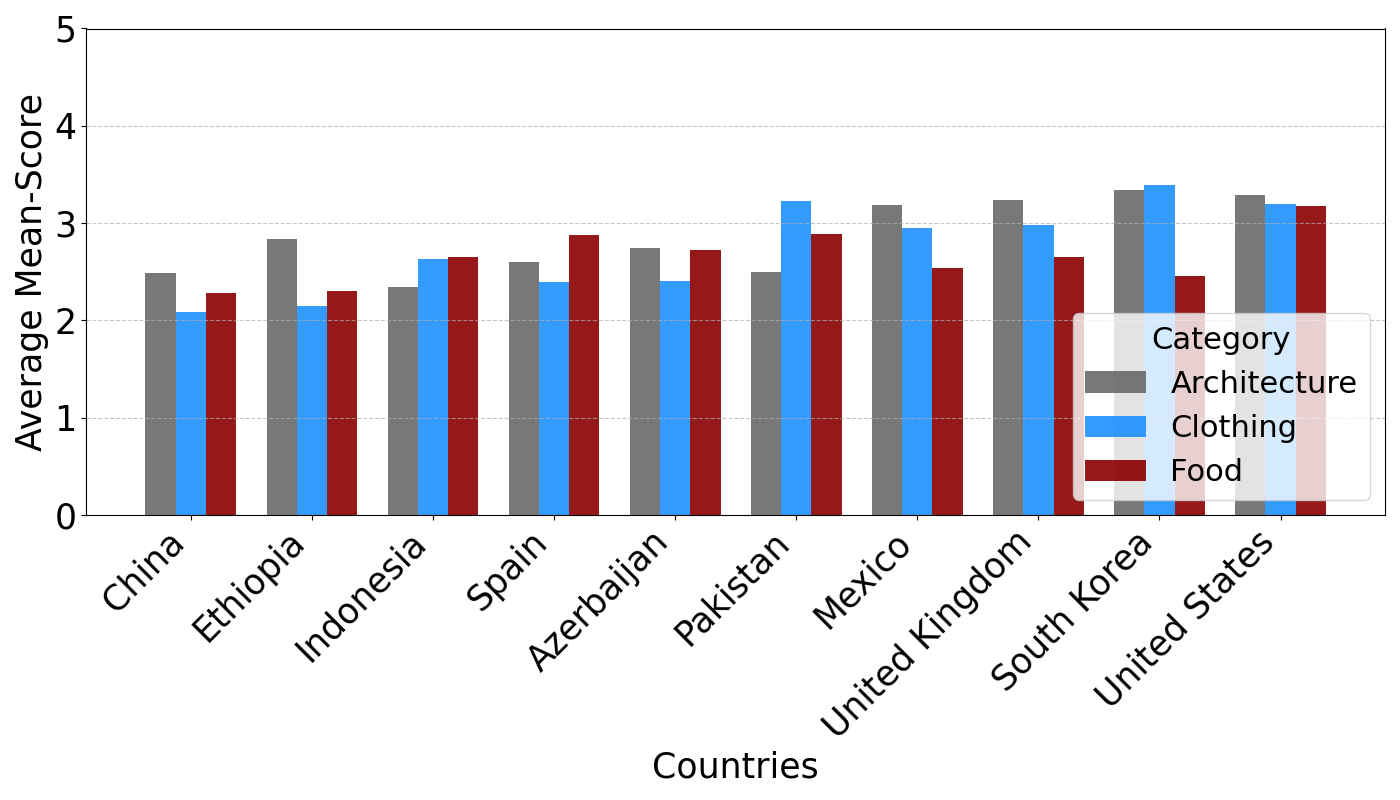}
  \includegraphics[width=0.32\textwidth]{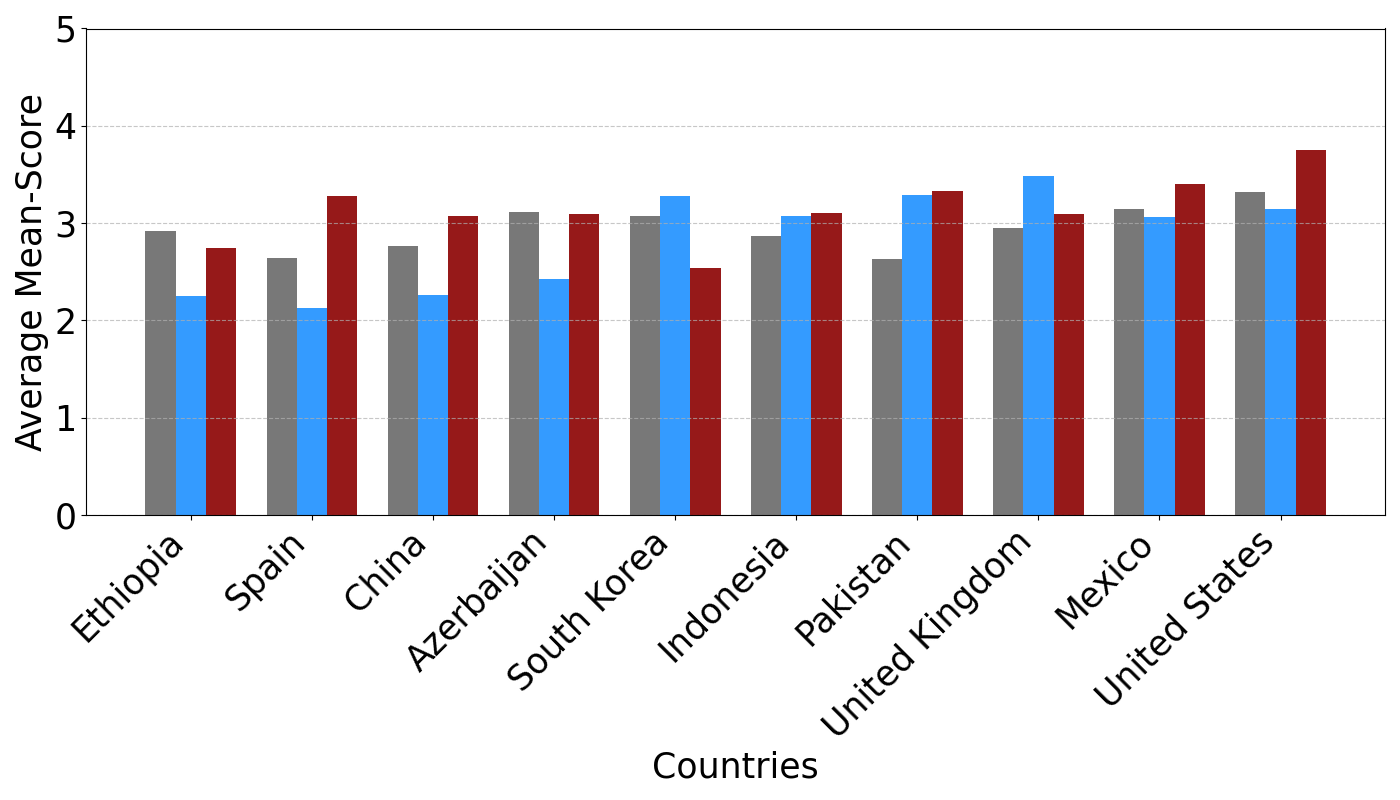}
  \includegraphics[width=0.32\textwidth]{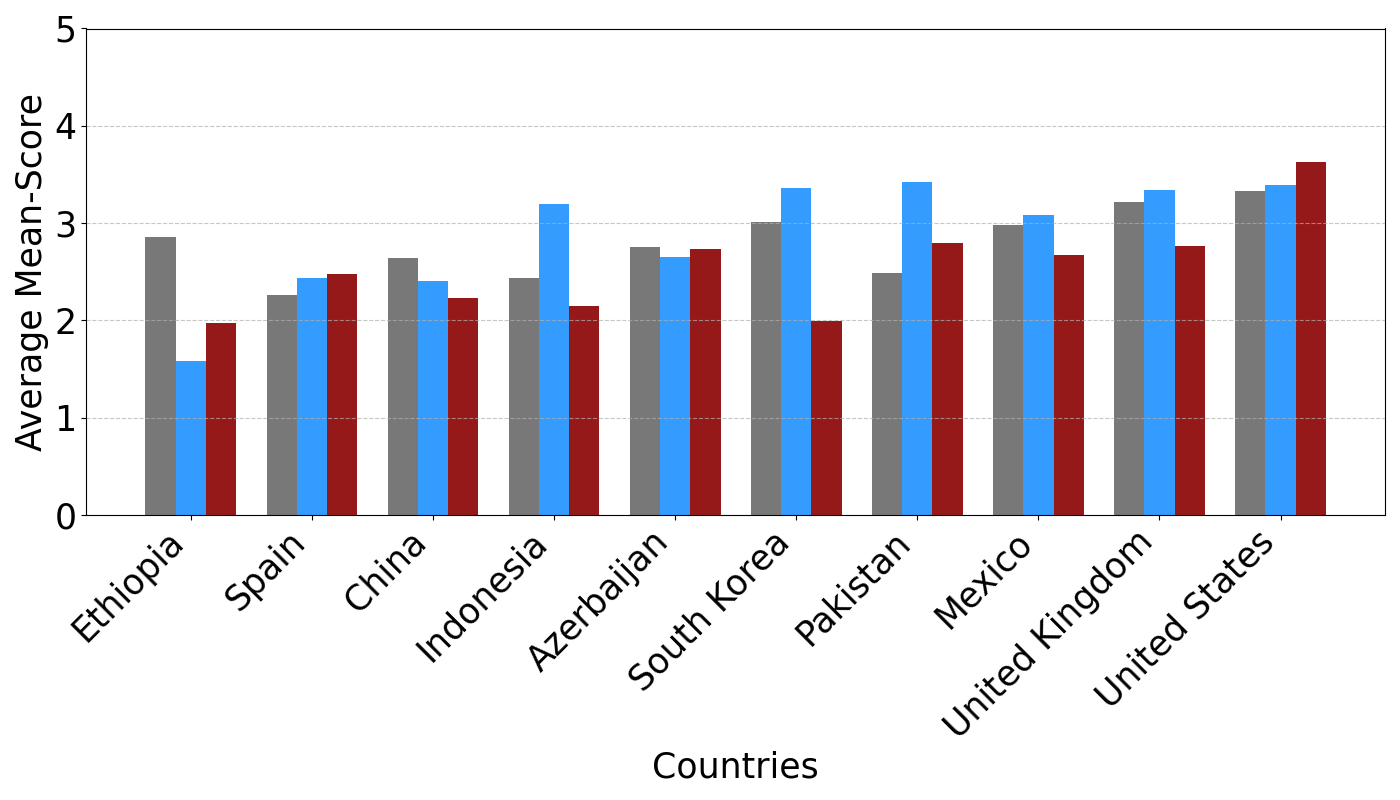}

  \caption{Average mean scores across categories for the outputs of three models: Stable-Diffusion XL (left), Stable-Diffusion 3 Medium Diffusers (center), and FLUX (right). The bar plots represent the average scores for Architecture, Clothing, and Food categories.}
  \label{fig:category_scores_comparison}
\end{figure*}

We analyzed the scores for each image quadruple by averaging the scores from six questions, assigning the resulting value as the final score for each image. We then calculated the average of these final scores for each category and country, as shown in Figure~\ref{fig:category_scores_comparison}.

\begin{figure*}[t]
  \centering
  \includegraphics[width=0.32\textwidth]{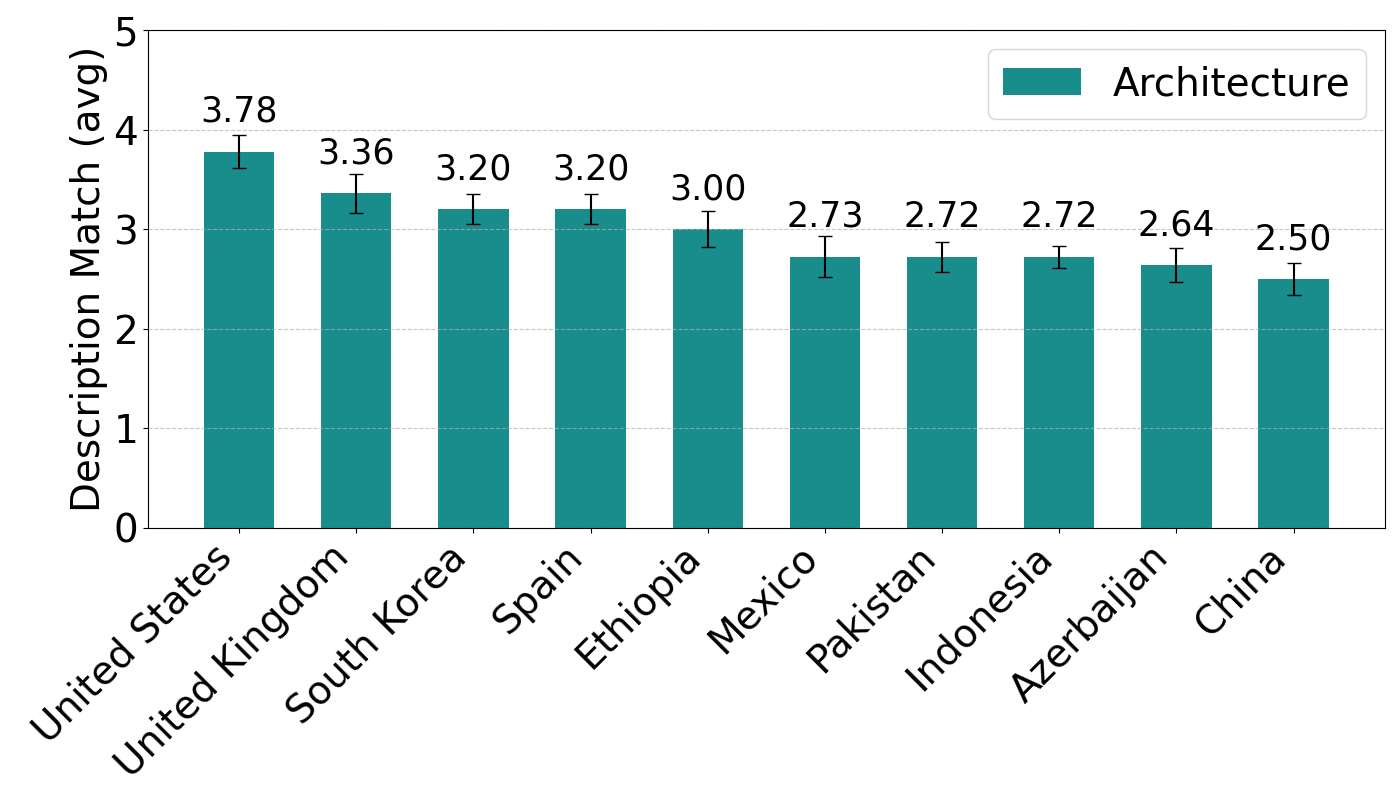}
  \includegraphics[width=0.32\textwidth]{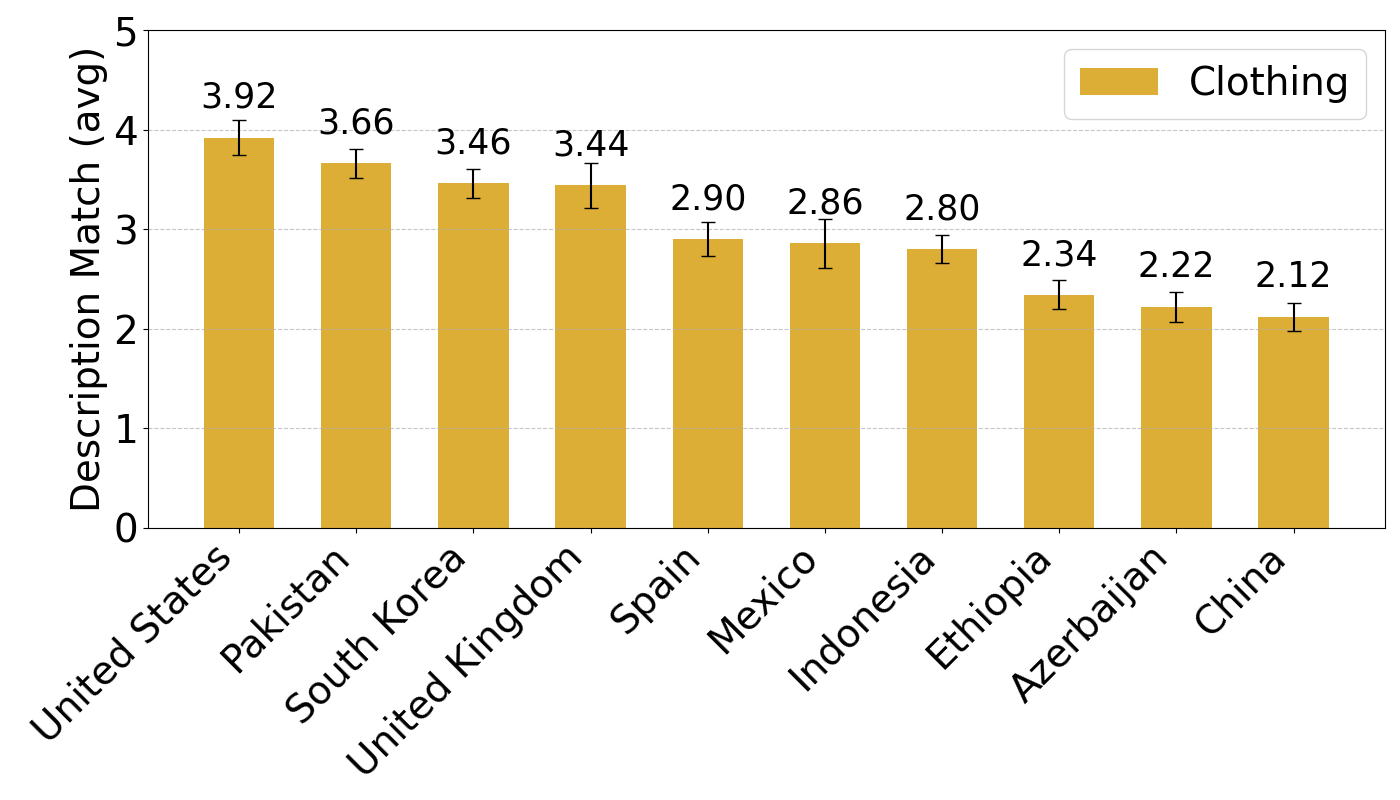}
  \includegraphics[width=0.32\textwidth]{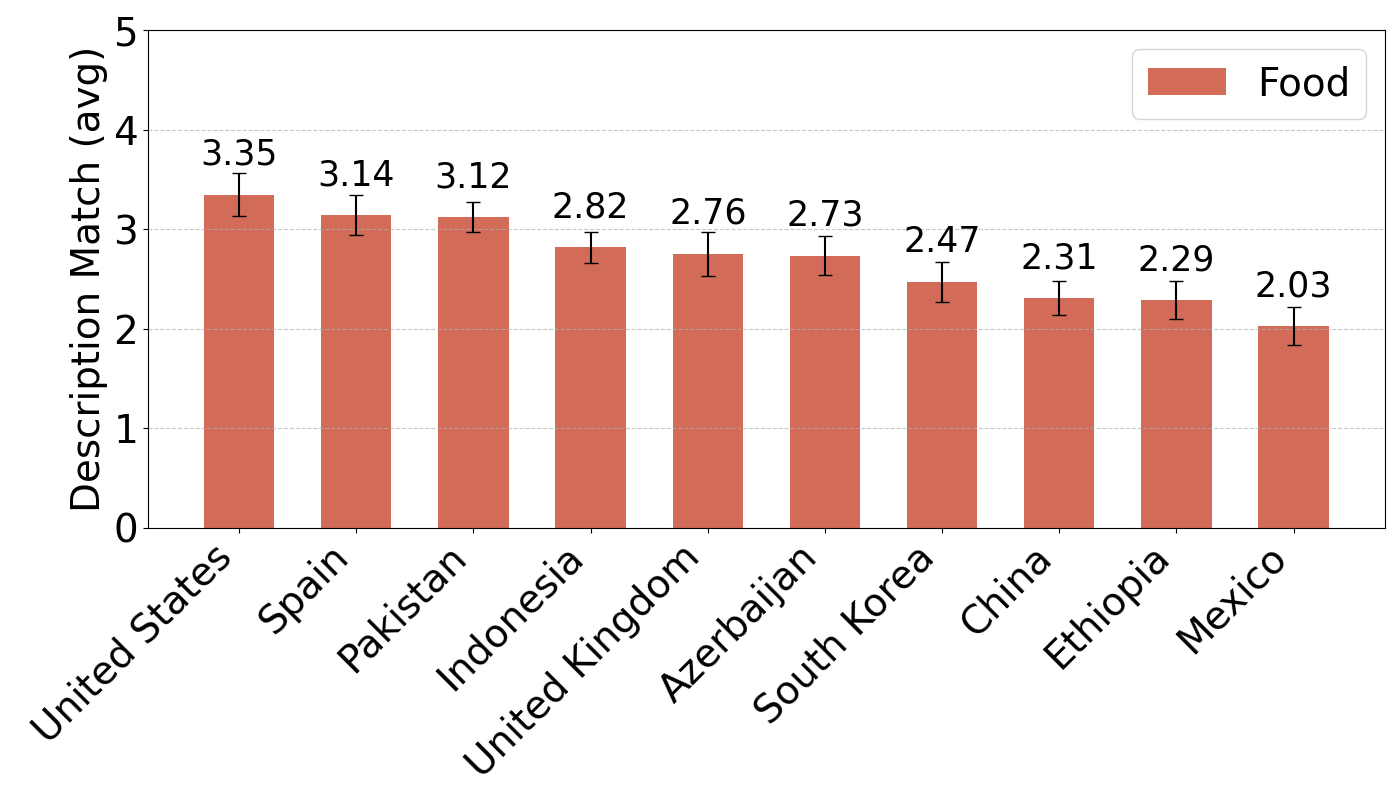}

  \caption{Average description match scores for the Stable-Diffusion XL model's outputs across three categories: Architecture (left), Clothing (center), and Food (right). Each bar plot represents the average match scores for countries within the respective category, ranked from highest to lowest. Error bars indicate the standard error of the mean.}
  \label{fig:description_match_xl}
\end{figure*}

\begin{figure*}[t]
  \centering
  \includegraphics[width=0.32\textwidth]{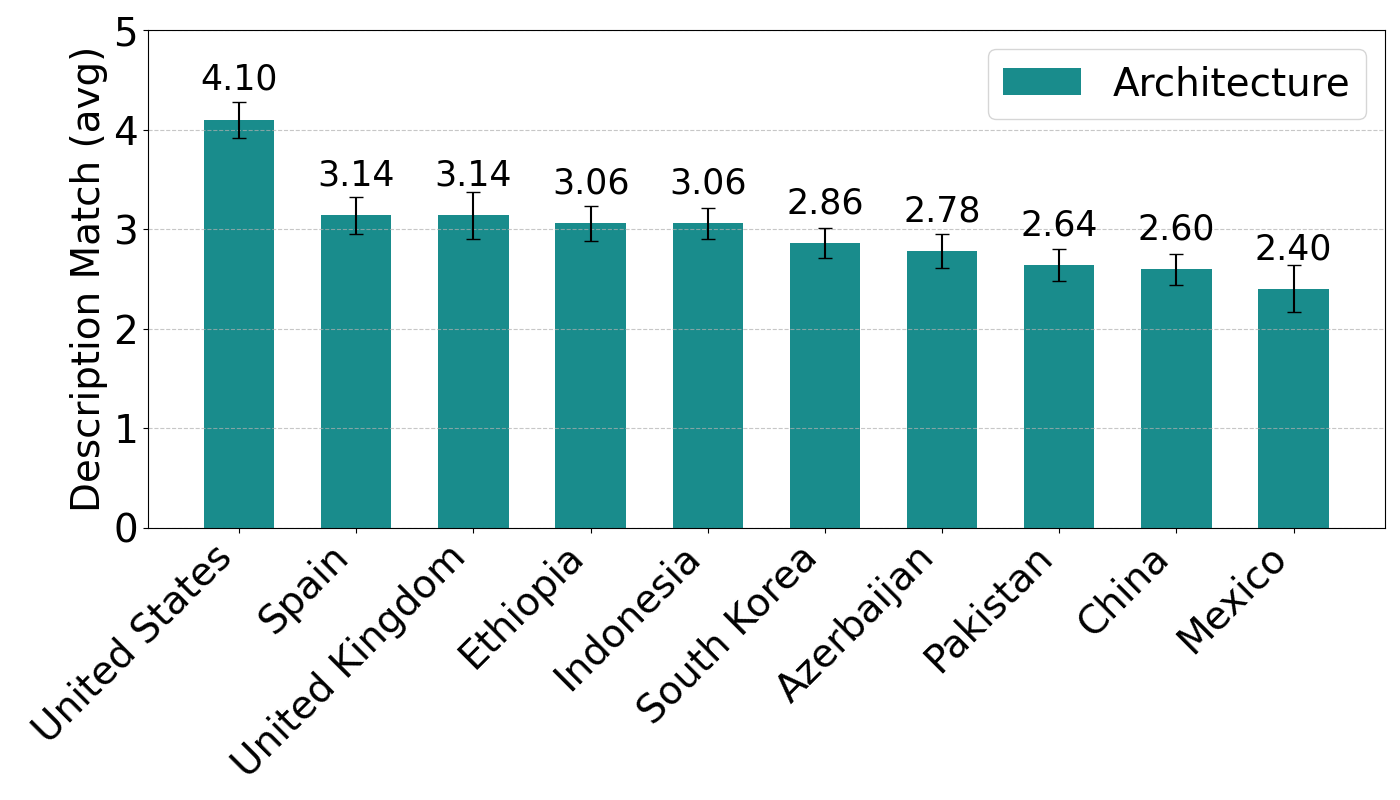}
  \includegraphics[width=0.32\textwidth]{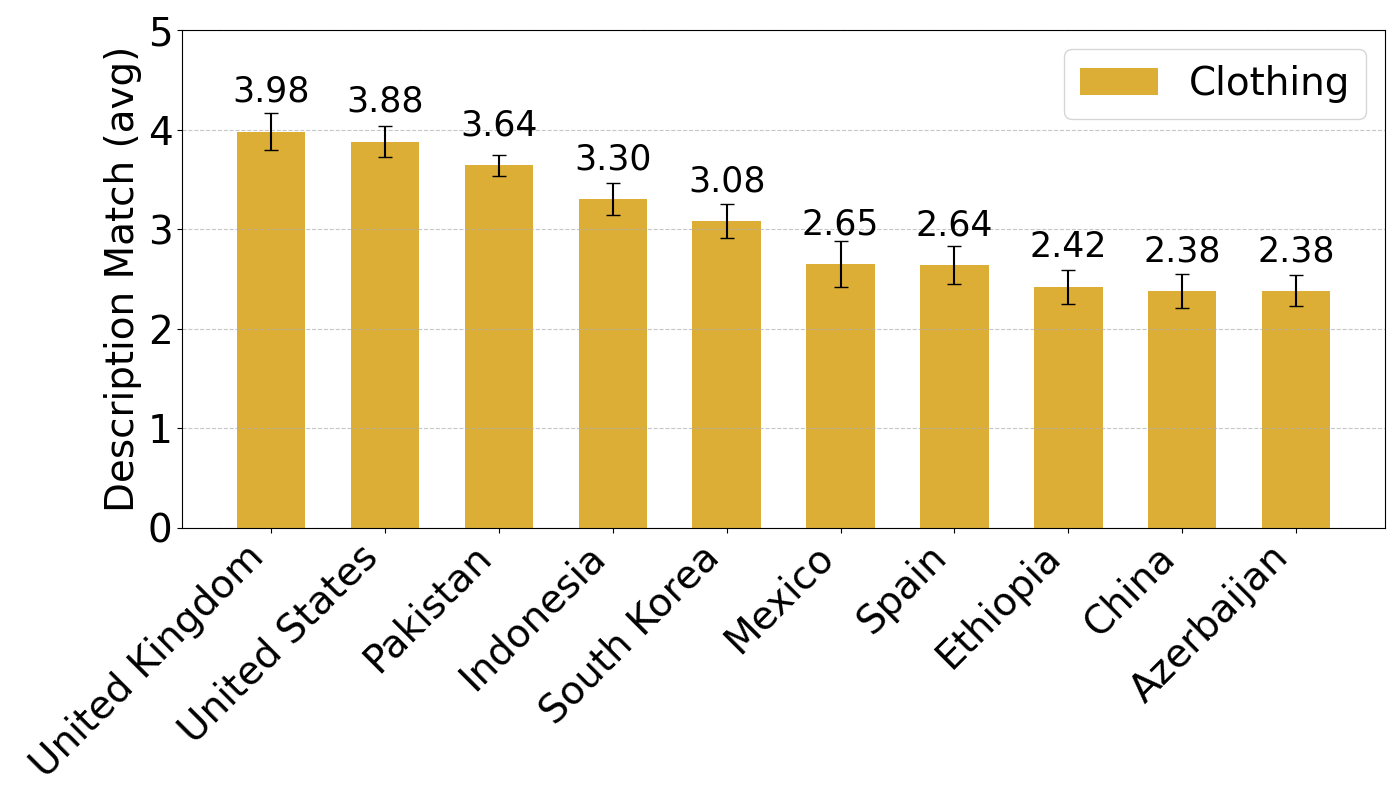}
  \includegraphics[width=0.32\textwidth]{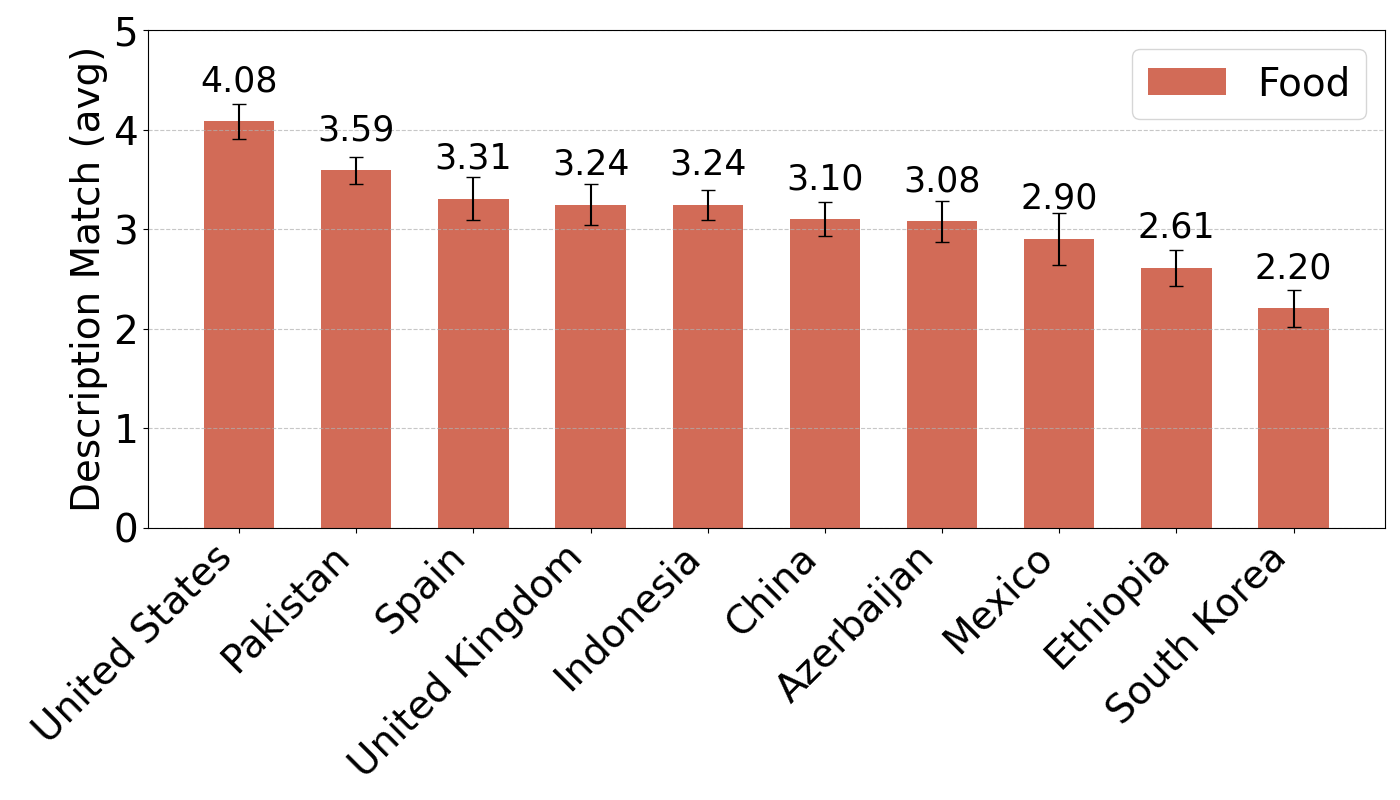}

  \caption{Average description match scores for the Stable-Diffusion 3 Medium Diffusers model's outputs across three categories: Architecture (left), Clothing (center), and Food (right). Each bar plot represents the average match scores for countries within the respective category, ranked from highest to lowest. Error bars indicate the standard error of the mean.}
  \label{fig:description_match_medium_diffusers}
\end{figure*}

The description match scores for Stable Diffusion XL and Stable-Diffusion 3 Medium Diffusers are presented in Figures~\ref{fig:description_match_xl} and~\ref{fig:description_match_medium_diffusers} respectively.

\subsection{Survey Response Correlations} \label{sec:appendix_a5}

\begin{figure*}[t]
  \centering
  \includegraphics[width=1.0\textwidth]{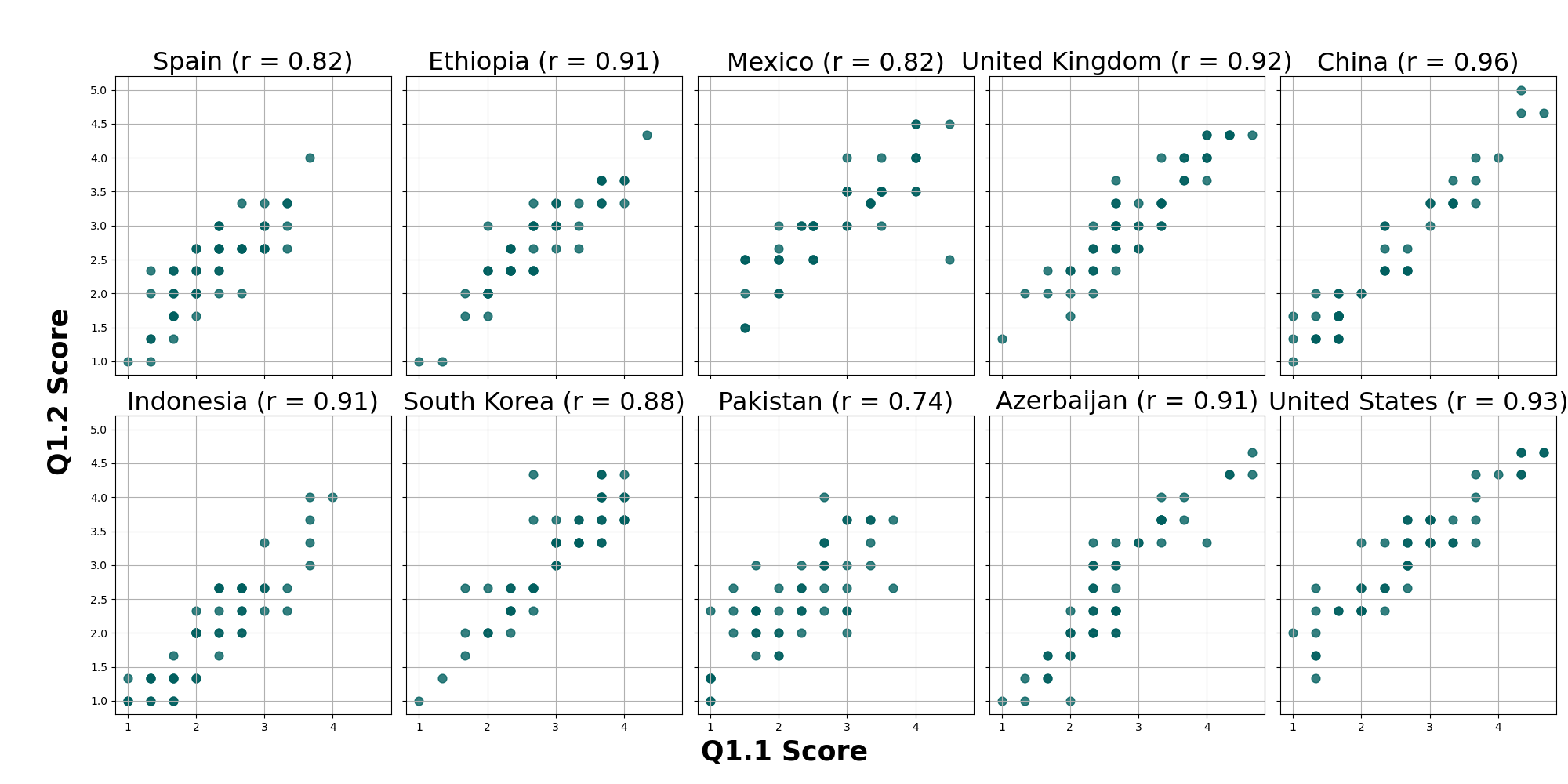}
  \caption{Scatter plots of Q1.1 (horizontal axis) vs. Q1.2 (vertical axis) scores for Architecture images, categorized by country. Each subplot displays the country name and corresponding Pearson correlation coefficient $r$. Dots represent individual image averages.}
  \label{fig:corr_q11_q12_architecture}
\end{figure*}

\begin{figure*}[t]
  \centering
  \includegraphics[width=1.0\textwidth]{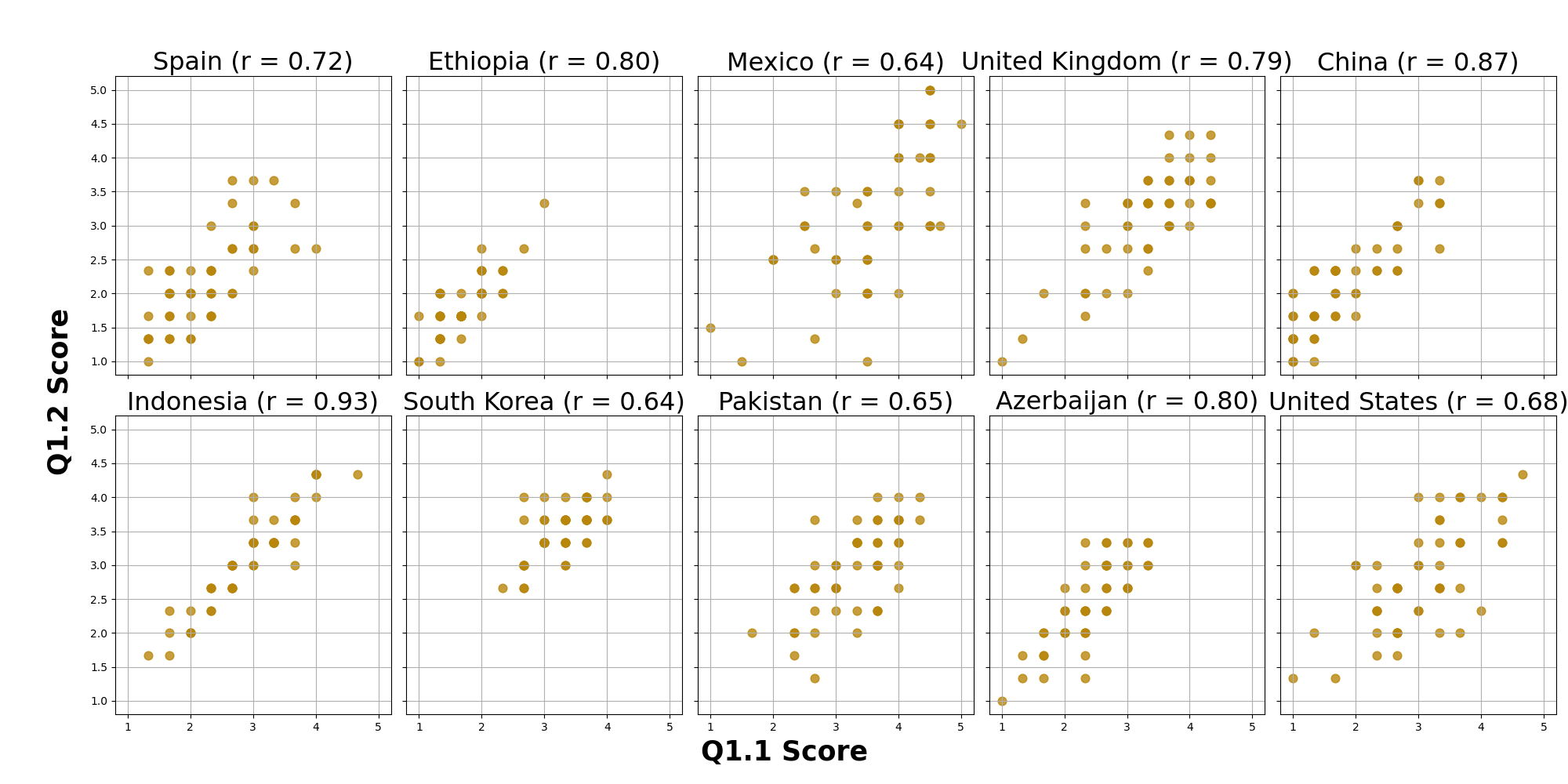}
  \caption{Scatter plots of Q1.1 (horizontal axis) vs. Q1.2 (vertical axis) scores for Clothing images, categorized by country. Each subplot displays the country name and corresponding Pearson correlation coefficient $r$.}
  \label{fig:corr_q11_q12_clothing}
\end{figure*}

\begin{figure*}[t]
  \centering
  \includegraphics[width=1.0\textwidth]{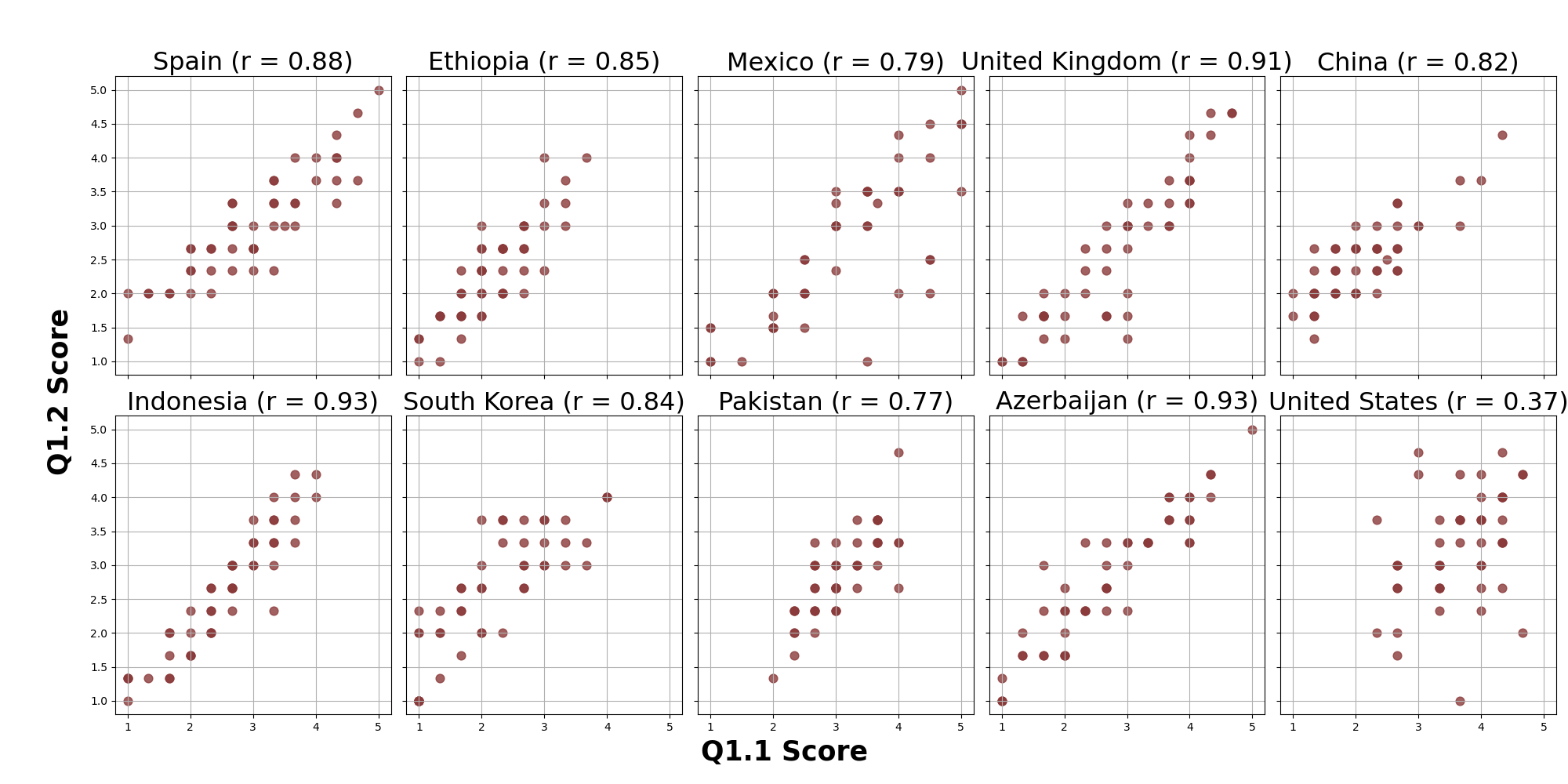}
  \caption{Scatter plots of Q1.1 (horizontal axis) vs. Q1.2 (vertical axis) scores for Food images, categorized by country. Each subplot displays the country name and corresponding Pearson correlation coefficient $r$.}
  \label{fig:corr_q11_q12_food}
\end{figure*}

\begin{figure*}[t]
  \centering
  \includegraphics[width=1.0\textwidth]{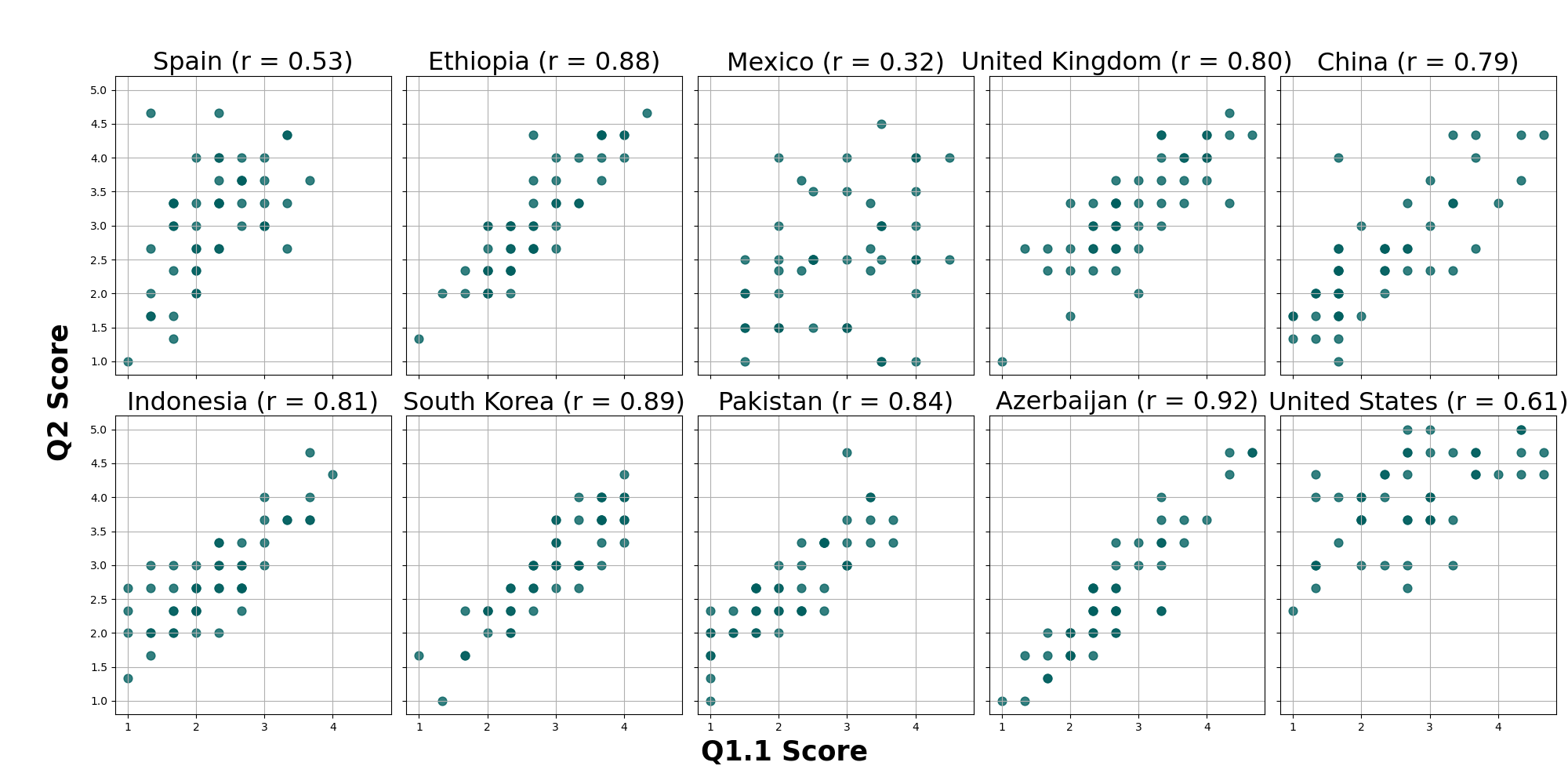}
  \caption{Scatter plots of Q1.1 (horizontal axis) vs. Q2 (vertical axis) scores for Architecture images, categorized by country. Each subplot displays the country name and corresponding Pearson correlation coefficient $r$.}
  \label{fig:corr_q11_q2_architecture}
\end{figure*}

\begin{figure*}[t]
  \centering
  \includegraphics[width=1.0\textwidth]{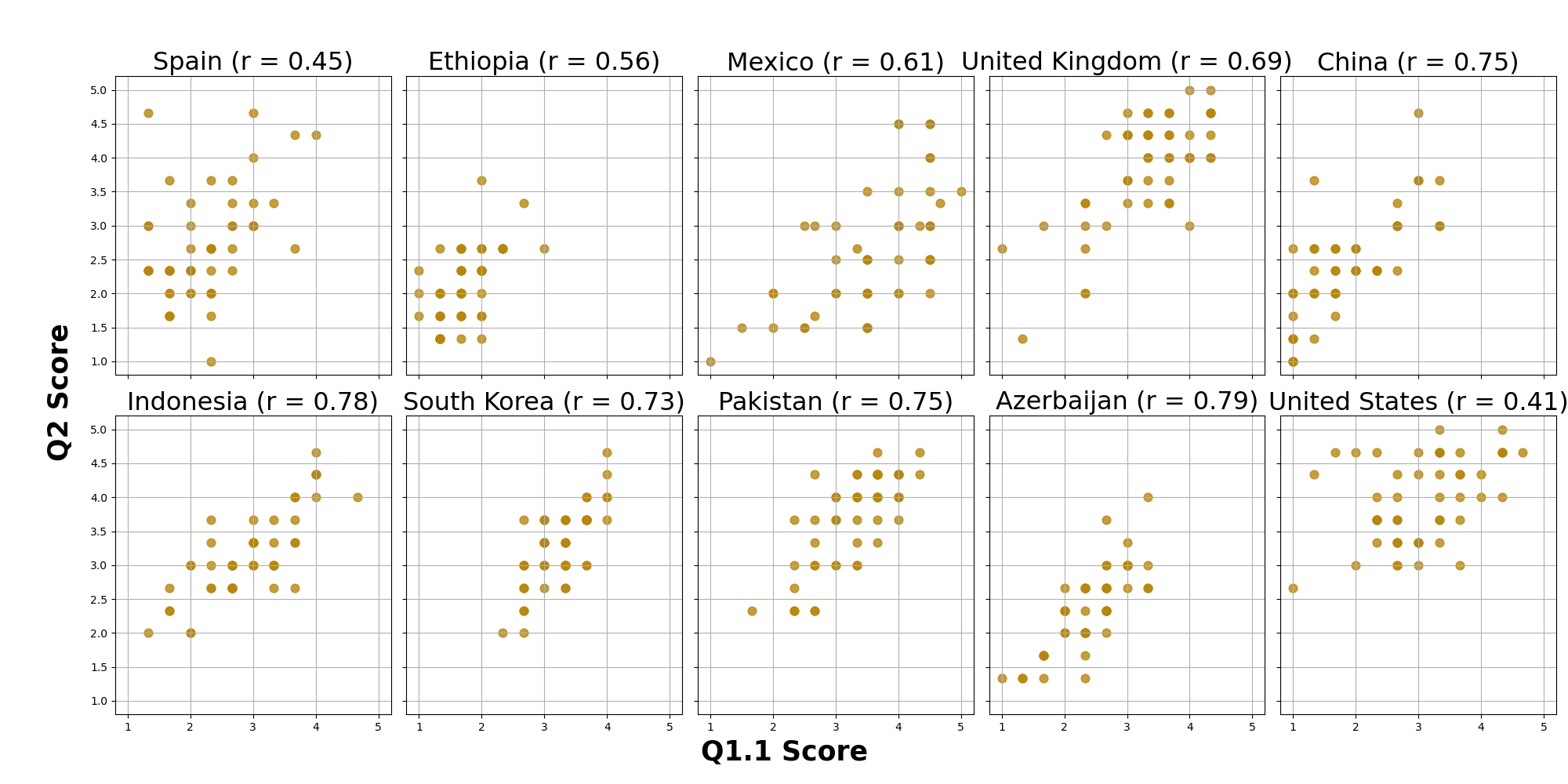}
  \caption{Scatter plots of Q1.1 (horizontal axis) vs. Q2 (vertical axis) scores for Clothing images, categorized by country. Each subplot displays the country name and corresponding Pearson correlation coefficient $r$.}
  \label{fig:corr_q11_q2_clothing}
\end{figure*}

\begin{figure*}[t]
  \centering
  \includegraphics[width=1.0\textwidth]{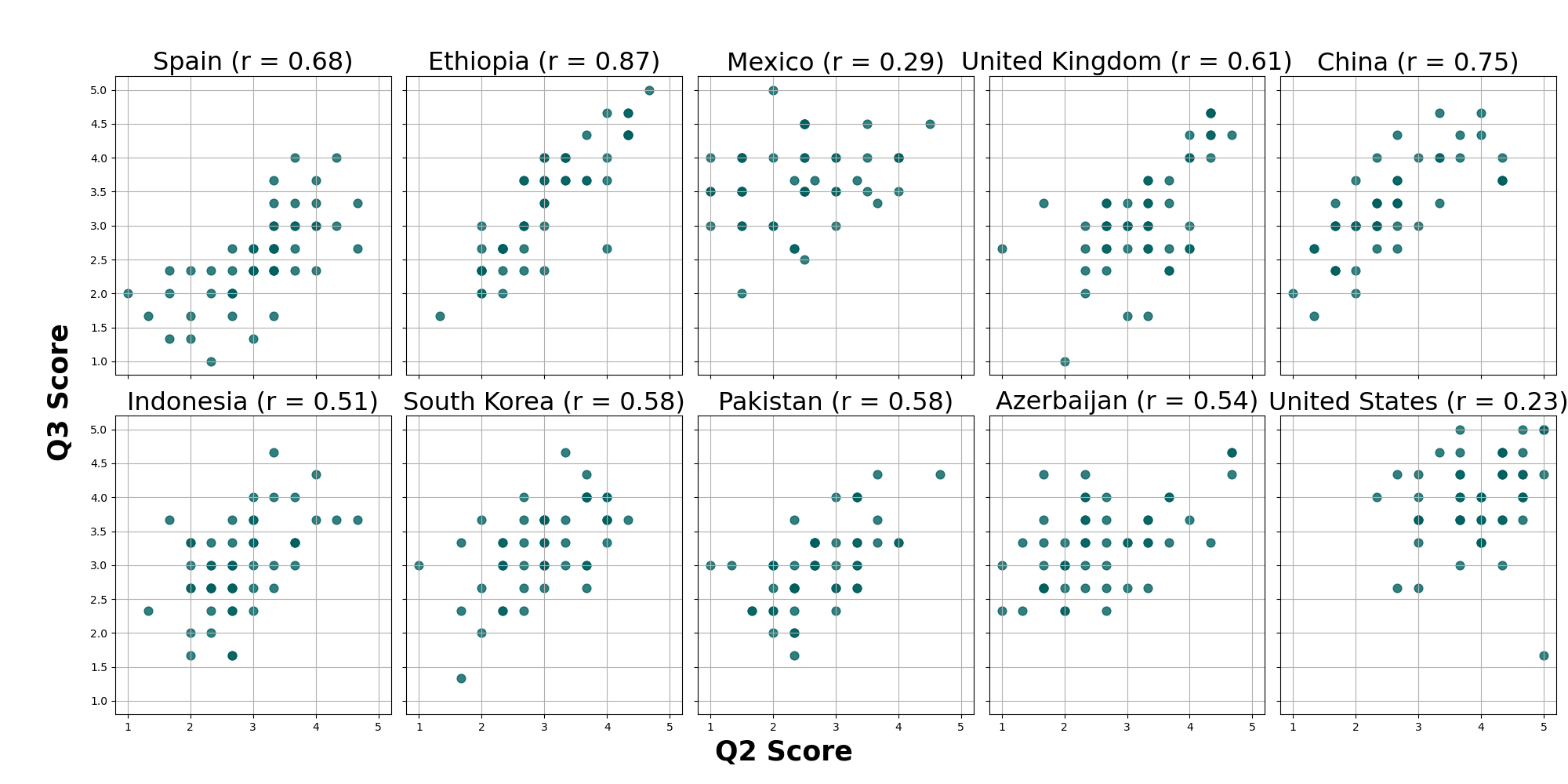}
  \caption{Scatter plots of Q2 (horizontal axis) vs. Q3 (vertical axis) scores for Architecture images, categorized by country. Each subplot displays the country name and corresponding Pearson correlation coefficient $r$.}
  \label{fig:corr_q2_q3_architecture}
\end{figure*}

\begin{figure*}[t]
  \centering
  \includegraphics[width=1.0\textwidth]{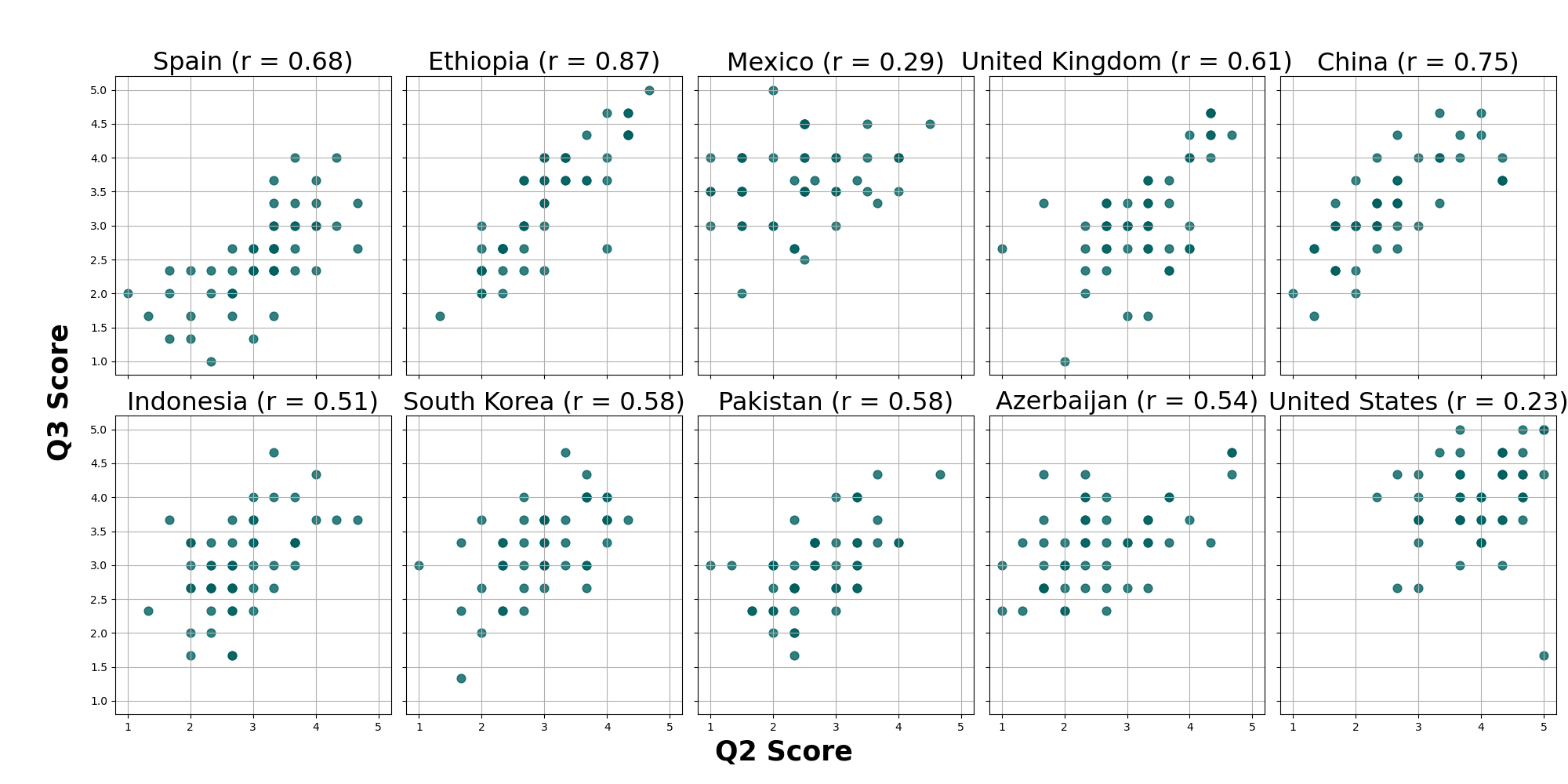}
  \caption{Scatter plots of Q2 (horizontal axis) vs. Q3 (vertical axis) scores for Clothing images, categorized by country. Each subplot displays the country name and corresponding Pearson correlation coefficient $r$.}
  \label{fig:corr_q2_q3_clothing}
\end{figure*}

\begin{figure*}[t]
  \centering
  \includegraphics[width=1.0\textwidth]{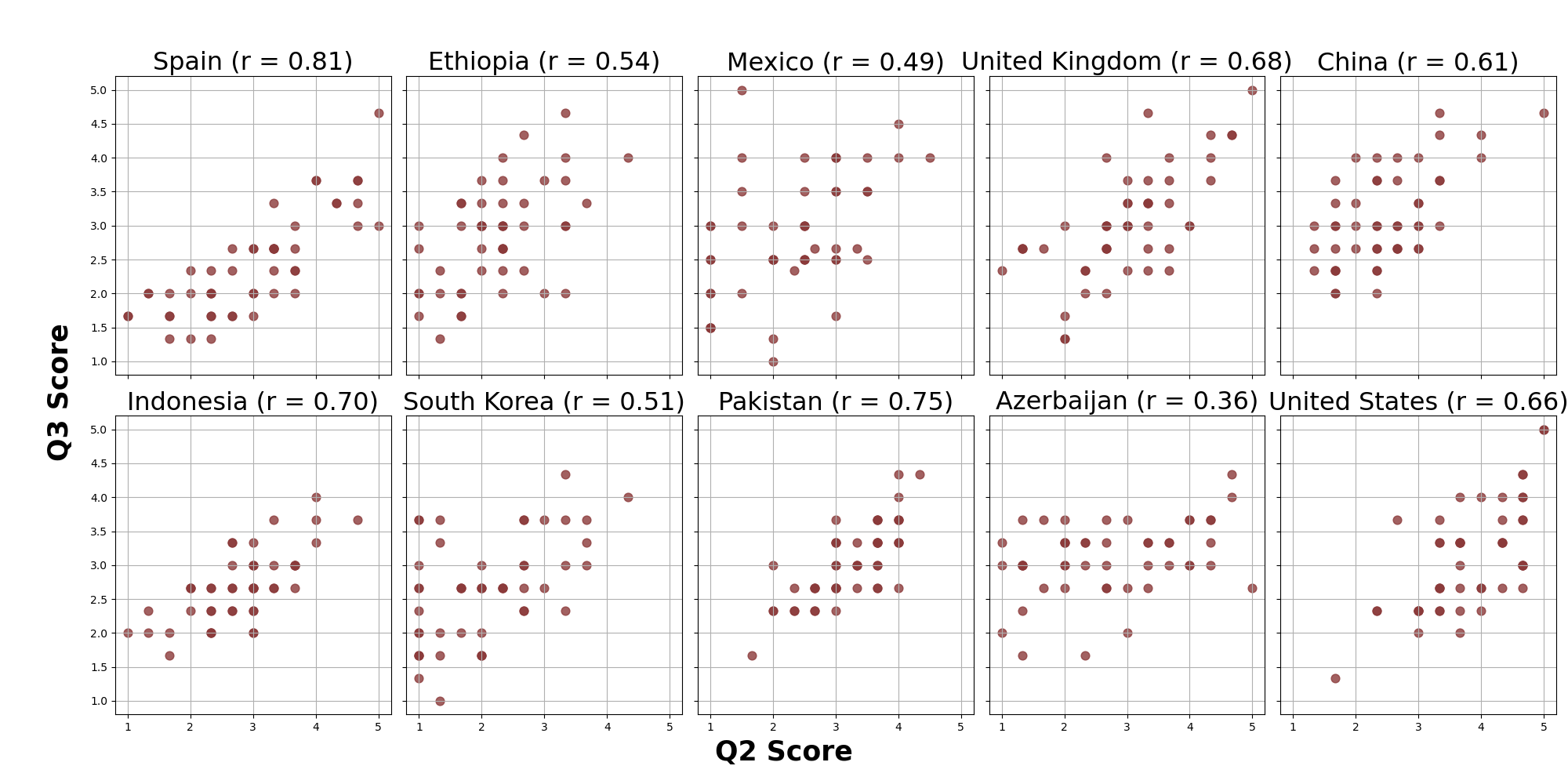}
  \caption{Scatter plots of Q2 (horizontal axis) vs. Q3 (vertical axis) scores for Food images, categorized by country. Each subplot displays the country name and corresponding Pearson correlation coefficient $r$.}
  \label{fig:corr_q2_q3_food}
\end{figure*}

We calculated the correlations between the scores for Q1.1 (overall similarity) and Q1.2 for three distinct image categories: Architecture, Clothing, and Food. For each image, we first averaged the responses across the models to compute a unified score. These scatter plots (Figure~\ref{fig:corr_q11_q12_architecture}, ~\ref{fig:corr_q11_q12_clothing}, ~\ref{fig:corr_q11_q12_food}) depict the relationship between the two survey questions.
Similarly, the correlation results for the Architecture and Clothing categories for Q1.1 (overall similarity) and Q2 (description match) are presented in Figures~\ref{fig:corr_q11_q2_architecture} and~\ref{fig:corr_q11_q2_clothing}, respectively. 
Moreover, we present the correlation scores for all three categories for Q2 (description match) and Q3 (realism) in Figures~\ref{fig:corr_q2_q3_architecture},~\ref{fig:corr_q2_q3_clothing}, and ~\ref{fig:corr_q2_q3_food}.

\subsection{Common Failure Patterns}
\label{sec:appendix_a6}
We provide a brief analysis of common failure patterns observed in the generated images. While our primary focus is on evaluating cultural accuracy, understanding these recurring errors offers insight into model limitations:
\begin{itemize}
    \item Azerbaijani Clothing: Models often produced garments resembling Central Asian (Uzbek, Kyrgyz) attire rather than accurately reflecting Azerbaijani traditional clothing.
    \item Ethiopian Architecture: Generated outputs frequently featured elements drawn from other African regions instead of distinctively Ethiopian structures, such as the rock-hewn churches of Lalibela.
    \item Korean Clothing: Models sometimes confused Korean traditional clothing with Chinese hanfu or Japanese kimono, reflecting difficulty in distinguishing nuanced East Asian cultural features.
    \item Pakistani Food: The models frequently generated Indian dishes such as dosa or idli when prompted for Pakistani cuisine.

\end{itemize}

\end{document}